%% file: main.tex
\documentclass[mnsc,nonblindrev]{informs3_nojournal}

\OneAndAHalfSpacedXI
\usepackage{amsmath,amssymb,amsfonts}
\usepackage[ruled]{algorithm2e}
\usepackage{bbm, dsfont}
\usepackage[hypertexnames=false]{hyperref}
\usepackage{xcolor,wrapfig}
\usepackage[capitalise]{cleveref}
\usepackage{subcaption}
\graphicspath{{./Figures/}}

\usepackage{graphicx}
\usepackage{multirow}
\usepackage{hhline}

\newcommand{\E}{\mathbb{E}}

\newcommand{\cX}{\mathcal{X}}

\newcommand{\cI}{\mathcal{I}}

\DeclareMathOperator{\opt}{OPT}

\input{OR/abbrv}

\usepackage{natbib}
 \bibpunct[, ]{(}{)}{,}{a}{}{,}%
 \def\bibfont{\small}%

\usepackage{rotating}
\usepackage{fancyvrb}

\TheoremsNumberedThrough     
\ECRepeatTheorems

\EquationsNumberedThrough    

\MANUSCRIPTNO{}

\begin{document}

\RUNAUTHOR{Cui, Jia and Lavastida}
\RUNTITLE{From Stream to Pool: Pricing Under the Law of Diminishing Marginal Utility}

\TITLE{From Stream to Pool: Pricing Under the Law of Diminishing Marginal Utility\footnote{Authors are alphabetically ordered} }

\ARTICLEAUTHORS{
\AUTHOR{Titing Cui \AFF{Joseph M. Katz Graduate School of Business, University of Pittsburgh} Su Jia \AFF{Center of Data Science for Enterprise and Society (CDSES), Cornell University} Thomas Lavastida \AFF{Naveen Jindal School of Management, UT Dallas}
}}

\ABSTRACT{Dynamic pricing models often posit that a \textbf{stream} of customer interactions occur sequentially, where customers' valuations are drawn independently.
However, this model is not entirely reflective of the real world, as it overlooks a critical aspect, the law of diminishing marginal utility, which states that a customer's marginal utility from each additional unit declines. 
This causes the valuation distribution to shift towards the lower end, which is not captured by the stream model.
This motivates us to study a pool-based model, where a \textbf{pool} of customers repeatedly interacts with a monopolist seller, each of whose valuation diminishes in the number of purchases made according to a discount function.
In particular, when the discount function is constant, our pool model recovers the stream model.
We focus on the most fundamental special case, where a customer's valuation becomes zero once a purchase is made.
Given $k$ prices, we present a non-adaptive, detail-free (i.e., does not ``know'' the valuations) policy that achieves a $1/k$ competitive ratio, which is optimal among non-adaptive policies.
Furthermore, based on a novel debiasing technique, we propose an adaptive learn-then-earn policy with a $\tilde O(k^{2/3} n^{2/3})$ regret.}


\maketitle

\section{Introduction}
\label{sec:intro}
\input{OR/Introduction}

\section{Formulation}\label{sec:model}
\input{OR/prelim}

\section{Detail-Dependent Policy}
\label{sec:known_demand}
\input{OR/known_demand}

\section{Non-Adaptive Detail-Free Policy}\label{sec:non_adap}
\input{OR/nonadap_V2}

\section{Adaptive Detail-Free Policy}\label{sec:adap}
\input{OR/Learning-New}

\section{Experiments}\label{sec:xp}
\input{OR/Numerics}

\section{Conclusion and Future Directions}\label{sec:future}
\input{OR/conclusion}

\input{OR/future}

\bibliographystyle{ormsv080}
\bibliography{OR/_pricing}

\begin{APPENDICES}
\input{OR/Appendix}
\end{APPENDICES}

\end{document}

%% file: OR/abbrv.tex
\newcommand{\new}{\color{black}}
\newcommand{\fff}{\color{black}} 

\newcommand{\su}[1]{\textcolor{red}{(SJ: #1)}}

\newcommand{\bitem}{\begin{itemize}}
\newcommand{\eitem}{\end{itemize}}
\newcommand{\benum}{\begin{enumerate}}
\newcommand{\eenum}{\end{enumerate}}
\newcommand{\bdefn}{\begin{definition}}
\newcommand{\edefn}{\end{definition}}
\newcommand{\bprop}{\begin{proposition}}
\newcommand{\eprop}{\end{proposition}}
\newcommand{\bque}{\begin{question}}
\newcommand{\eque}{\end{question}}
\newcommand{\bobsv}{\begin{observation}}
\newcommand{\eobsv}{\end{observation}}
\newcommand{\beqn}{\begin{equation}\begin{aligned}}
\newcommand{\eeqn}{\end{aligned}\end{equation}}

\newcommand{\ps}{\begin{proof}[Sketch]}
\newcommand{\brmk}{\begin{remark}}
\newcommand{\ermk}{\end{remark}}
\newcommand{\bduiqi}{\begin{aligned}}
\newcommand{\eduiqi}{\end{aligned}}
\newcommand{\bcoro}{\begin{corollary}}
\newcommand{\ecoro}{\end{corollary}}
\newcommand{\bcom}{}

\newcommand{\adap}{adaptive}

\newcommand{\apxn}{approximation}

\newcommand{\arb}{arbitrary}

\newcommand{\alg}{algorithm}

\newcommand{\cvx}{convex}

\newcommand{\compet}{competitive}

\newcommand{\conti}{continuous}

\newcommand{\ci}{confidence interval}

\newcommand{\distr}{distribution}

\newcommand{\hypo}{hypothesis}

\newcommand{\ho}{\mathbb}

\newcommand{\indep}{independent}

\newcommand{\ih}{induction hypothesis}

\newcommand{\ins}{instance}

\newcommand{\inv}{inventory}

\newcommand{\ineq}{inequality}

\newcommand{\kehua}{characterize}

\newcommand{\lam}{\lambda}
\newcommand{\lan}{\langle}

\newcommand{\mtg}{martingale}

\newcommand{\mtxs}{matrices}

\newcommand{\Omg}{\Omega}

\newcommand{\obj}{objective}

\newcommand{\rb}{\right}
\newcommand{\lb}{\left}

\newcommand{\prb}{probability}

\newcommand{\parti}{particular}

\newcommand{\pmt}{parameter}

\newcommand{\rv}{random variable}

\newcommand{\rar}{\rightarrow}

\newcommand{\ran}{\rangle}

\newcommand{\real}{\mathbb{R}}
\newcommand{\resp}{respectively}

\newcommand{\sats}{satisfies}

\newcommand{\sce}{scenario}

\newcommand{\sln}{solution}
\newcommand{\sse}{\subseteq}
\newcommand{\sps}{suppose}
\newcommand{\Sps}{Suppose}

\newcommand{\strfwd}{straightforward}

\newcommand{\stoch}{stochastic}

\newcommand{\sym}{symmetry}

\newcommand{\unif}{uniform}

\newcommand{\wh}{\widehat}

\newcommand{\xulie}{sequence}

\let\eps\varepsilon

%% file: OR/Introduction.tex
Pricing with an unknown demand function is a fundamental challenge in revenue management.
Originating from the seminal work of \citet{gallego1994optimal}, many existing works employ what we call the {\bf stream} model:
a stream of customers arrives sequentially, wherein each customer makes a purchase if the current price is lower than their valuation, which is an i.i.d. draw from a fixed distribution.
The seller aims to maximize the total revenue by controlling the price. 

However, the stream model does not adequately model many real-world scenarios. 
Consider the sale of new clothing items.
Customers sporadically monitor the price (by visiting the store in person or checking the price online) based on their availability. 
A customer usually needs at most one unit of the clothing, so they exit the market immediately after making a purchase. 
In other words, their valuation has {\em diminished} to zero.
Moreover, due to the product's affordability, customers do not consider it worthwhile to strategically time their purchases, even if they are aware that the price might decrease.

In this example, the demands are {\bf neither} identically nor independently distributed. 
On the one hand, high-valuation customers tend to purchase early and subsequently leave the market, resulting in a {\em shift} in the valuation distribution towards the lower end.
Although non-stationary demand models have been studied (see, e.g., \citealt{besbes2011minimax, besbes2014dynamic, den2015tracking}), the non-stationarity is often {\bf exogenous}, incorporating external factors such as seasonality instead of the seller's actions.
In contrast, the non-stationarity in the clothing example is {\bf endogenous}, governed by the seller's decisions. 
Specifically, the price sequence affects how the remaining customers' valuations are distributed.

Orthogonal to non-stationarity, the independence assumption is also questionable in the above scenario. 
Consider a two-period {\em markdown} policy that reduces the price from high to low, and a {\em markup} policy that chooses these two prices in reverse order.
In the stream model, these two policies have the same expected revenue. 
However, in the above example the markup policy yields less revenue than the markdown policy because high-valuation customers might take advantage of the low initial price, and thus the seller misses out on higher revenue opportunities. 

\subsection{The Pool Model: An Informal Description}
The above discussion motivates us to introduce the following {\em pool model} (PM).
A {\bf pool} of myopic customers is present at the beginning of the time horizon. 
Each customer {\em interacts} with the seller by peeking at the price according to an independent Poisson process (whose parameter we refer to as the \emph{interaction rate}).
When such an interaction occurs, the customer purchases one unit if her valuation is higher than the price.

The valuation of each customer is proportional to a {\em discount function} $\psi(m)$ where $m$ is the number of purchases she has made. 
This function encompasses a fundamental aspect often overlooked in the study of dynamic pricing: The marginal utility of each additional purchase decreases. 
This phenomenon is commonly known as the {\em law of diminishing marginal utility} \citep{gossen1854entwickelung} and is well studied in economics; see, e.g., \citealt{kahneman1979prospect,arrow1974essays}, and \citealt{rabin2000risk}.

One facet of this problem, where $\psi(m)$ is constant, has been well understood. 
In fact, the pool model is {\bf equivalent} to the stream model (see Theorem \ref{thm:unify_pool_and_stream}) in this case, which has been extensively studied 
\citep{kleinberg2003value,besbes2009dynamic,babaioff2015dynamic}.
For example, if the unknown valuation distribution is supported on a known set of cardinality $k$, the stream model is equivalent to $k$-armed bandits, which has an  $\widetilde O(\sqrt{kn})$ minimax regret \citep{auer2002finite}.
(Here, the notation $\tilde O$ suppresses poly-logarithmic terms in $n,k$.)

However, little is known about the opposite (albeit equally fundamental) extreme where $\psi(m) = \mathbbm{1}(m=0)$, which we call as the {\em unit-demand pool model} (UDPM).
In this setting, each customer exits the market permanently once a purchase is made. 
This scenario has only been previously analyzed in the {\em immediate-buy} setting where the interaction rate is infinite \citep{talluri2006theory}, which is a somewhat straightforward scenario; see our discussion in \cref{sec:known_demand}.

\begin{wrapfigure}{R}{8.5cm}
\centering
\includegraphics[width=8.5cm, height=5cm]{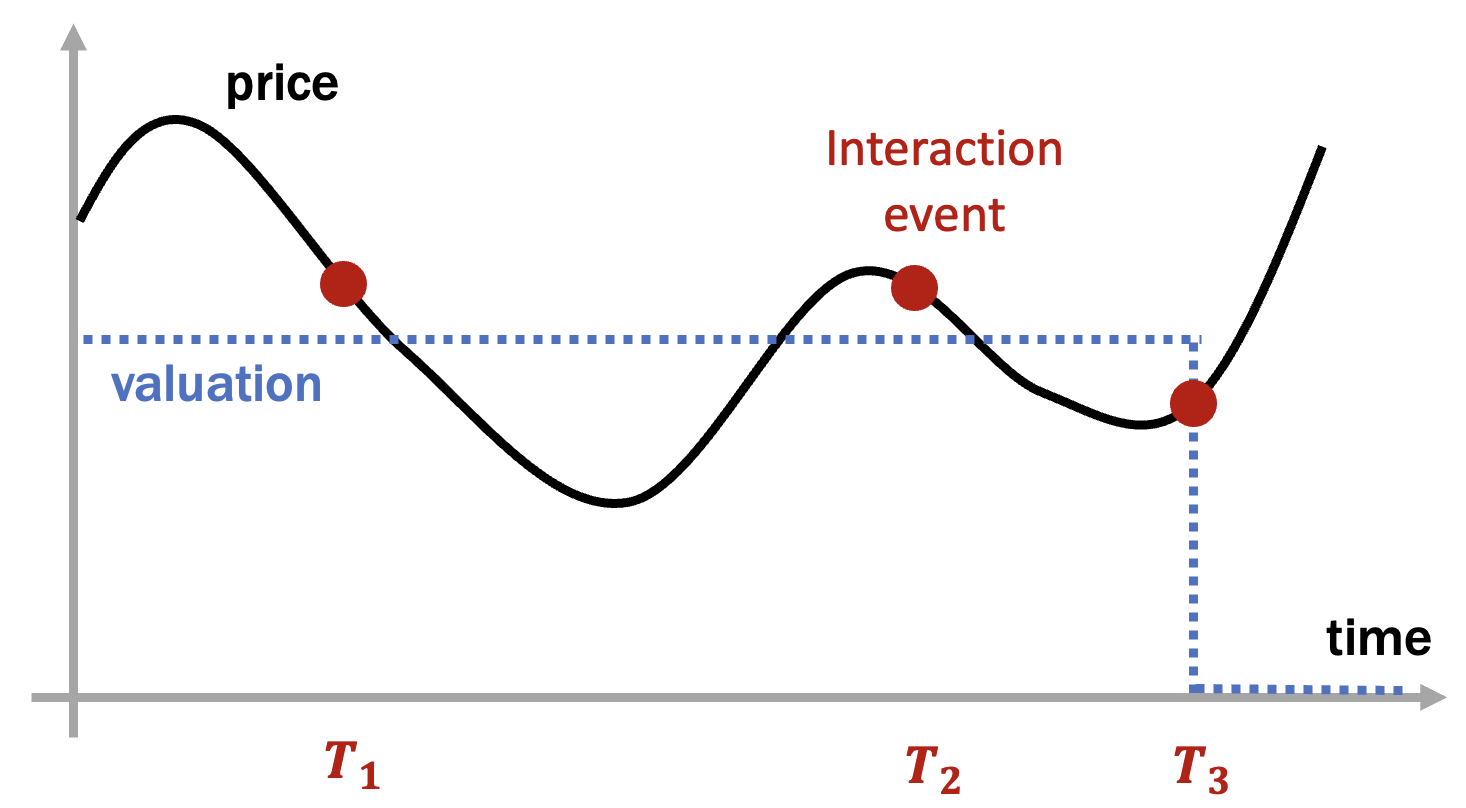}
\caption{Illustration of the UDPM: Red dots represent customer interactions, where they peek at the price. 
A purchase is made at the first interaction event ($T_3$) where the price is lower than the valuation.}
\label{fig:UDPM}
\end{wrapfigure}

We emphasize that in our model, customers make myopic decisions
based only on the current price.
The myopic model provides a reasonable approximation of customer behavior in many scenarios, particularly where the unit price is relatively low, making strategic behavior unappealing to most customers.
As \citet{li2014consumers} observed from air travel data sets, (only) ``5.2\% to 19.2\% of the population are strategic''.

Against this backdrop, we study the pool model, with a focus on the most basic setting - the UDPM.
We first show that when the valuations are known, 
we can handle the problem using fairly standard techniques.
Then, we present non-adaptive and adaptive policies that do not require knowledge of the valuations and yet exhibit robust performance.





\subsection{Our Contributions}
\begin{table}[]
\centering
\begin{tabular}{lllll}
\hline
\multicolumn{1}{|c|}{Model} & \multicolumn{1}{c|}{Discount Function} & \multicolumn{1}{c|}{Optimal Non-adaptive Policy} & \multicolumn{1}{c|}{Optimal CR} 
& \multicolumn{1}{c|}{Minimax Regret} \\ \hline
\multicolumn{1}{|c|}{Stream}  & \multicolumn{1}{c|}{$\psi(m)\equiv 1$} & \multicolumn{1}{c|}{fixed price policy} & \multicolumn{1}{c|}{$1/k$} 
& \multicolumn{1}{c|}{$\widetilde \Theta\lb(k^{1/2} n^{1/2}\rb)$} \\ \hline
\multicolumn{1}{|c|}{UDPM} & \multicolumn{1}{c|}{$\psi(m)=\mathbbm{1}(m=0)$} & \multicolumn{1}{c|}{{\color{blue} markdown policy}}   & \multicolumn{1}{c|}{{\color{blue} $1/k$}} & \multicolumn{1}{c|}{ {\color{blue}$\widetilde O\lb(k^{2/3} n^{2/3}\rb)$}}\\ \hline 
\end{tabular}
\caption{Positioning of our results.
``CR'' and ``Minimax regret'' refer to the best possible competitive ratio and regret achievable by a detail-free non-adaptive and adaptive policy.
Our results are colored red.
}
\end{table}

We initiate the study of the pool model, focusing on the simplest yet fundamental setting where $\psi(m)=\mathbbm{1}(m=0)$. 
We contribute to the study of dynamic pricing in the following ways. 

\noindent{\bf 1. A Novel Model: Unification of Two Worlds.} 
Our primary contribution is the introduction of a novel pool-based pricing model, which better aligns with reality in many scenarios compared to the extensively studied stream model.
Our pool model also encompasses the stream model as a special case (see \cref{thm:unify_pool_and_stream}), thereby broadening the scope of dynamic pricing and providing new avenues for further research.

\noindent{\bf 2. Known Valuations: Computing Optimal Policies.}
We provide two algorithms for computing a near-optimal policy, assuming that the valuations are known.
As the structure of the optimal adaptive policy is intricate, we will focus on non-adaptive policies (i.e., with predetermined price sequences). 
Specifically, we show the following:
\benum 
\item[(a)] {\bf Price Monotonicity.} We show that there exists an optimal non-adaptive policy whose price sequence is non-increasing; see Proposition \ref{prop:monotone}.
These policies are usually called {\em markdown policies}.
 
\item[(b)] {\bf Concavity of the Revenue Function.} (a) suggests that we may restrict our attention to markdown policies, which can be represented as $k$-dimension vectors.
We show that the expected revenue is a concave function in the markdown policy; see \cref{prop:concave}.

\item[(c)] {\bf Computing an Optimal Non-Adaptive Policy.} Building upon (a) and (b), we present two efficient \alg s for computing an optimal non-adaptive policy, based on dynamic programming and concave maximization.
\eenum 

\noindent{\bf 3. Unknown Valuations: Competitive Non-Adaptive Policies.} 
We present detail-free (i.e., do not ``know'' the valuations) policies that enjoy robust performance against any policy, possibly adaptive or detail-dependent (i.e., knows the valuations).
We provide a complete settlement by showing the following. 
\benum
\item[(a)] {\bf Competitive Policy for Finite Price Set.}
Given any set of $k$ prices, we present a non-\adap, detail-free policy that guarantees a $(1/k)$ fraction of the revenue achievable by any policy; see \cref{thm:non-adap}. 
\item[(b)] { {\bf Competitive Policy for Infinite Price Set.} Given any (continuous) price interval $[p_{\rm min}, p_{\rm max}]$, we construct a detail-free, non-adaptive policy which guarantees a $1/(1+\ln \rho)$ fraction of the optimum, where $\rho = p_{\rm min} / p_{\rm max}$; see \cref{thm:conti_P}.}

{ \item[(c)] {\bf Optimality.} Our \alg\ achieves the optimal \compet\ ratio. 
Specifically, we show that for any $\eps>0$ and any non-adaptive policy, there is a $k$-price instance on which this policy yields no more than $1/k + \eps$ fraction of the optimal revenue; see \cref{thm:non_adap_upper}.}  
A similar result can be shown for price intervals.
\eenum

\noindent{\bf 4. Unknown Valuations: Adaptive Policy with Sublinear Regret.}
Despite their simplicity, non-adaptive policies have two major drawbacks: Their performance (i) does not improve as $n$ increases, and (ii) deteriorates {\em linearly} in $k$.
Thus, we study adaptive policies, presenting one with a $\tilde O(k^{2/3} n^{2/3})$ regret against the optimal non-adaptive policy; see \cref{thm:regret_bound}. 
This is achieved by combining the following components. 
\benum
\item[(a)] {\bf Debiasing the Demand.} Unlike the stream model, which is equivalent to {\em multi-armed bandits} (MAB), we face an additional challenge of {\em confounding} observations.
Specifically, when we select a price $p$, customers with valuations greater than $p$ can also buy, but we do not observe the customers' valuations.
This makes it difficult to estimate the valuations.
To address this, we propose a novel unbiased estimator for the valuations.

\item[(b)] {\bf Achieving $o(n)$ Regret.} 
We propose a simple {\em learn-then-earn} policy based on the above estimator. 
When $k=2$, this policy achieves a $\tilde O(n^{2/3})$ regret as $n\rar \infty$.
A \strfwd\ extension of the analysis leads to a  $\tilde O(kn^{2/3})$ regret bound for \arb\ $k<\infty$.

\item[(c)] {\bf Achieving $o(kn)$ Regret.} 
{\new Unlike in MAB, it is not \strfwd\ to obtain regret bounds that are sublinear in $k$.}
We improve the dependence on $k$ via a more refined analysis.
We show that the cumulative estimation error prices form a {\em self-neutralizing} process, which, loosely, means that the estimation error at the $j$-th price level has the tendency to cancel the cumulative (signed) errors over the previous $(j-1)$ price levels.
Furthermore, we show that the sequence of estimation errors satisfies {\em conditional sub-gaussianity}.
These two properties enable us to obtain the claimed regret bound. 
\eenum


\subsection{Related Work}\label{sec:lit_rev}
Our work is related to the following lines of research.

\subsubsection{Dynamic Pricing in the Stream Model}
The stream model has been extensively studied since the seminal work of \citet{gallego1994optimal}.
The problem is particularly intriguing when the demand model is unknown, where the seller must balance learning and earning
\citep{kleinberg2003value}. 
Various fundamental aspects have also been investigated, including finite inventory \citep{besbes2009dynamic, wang2014close,babaioff2015dynamic,den2015dynamic1}, parametric demand functions \citep{den2013simultaneously}, 
joint inventory-pricing control \citep{chen2004coordinating,chen2019coordinating}, customer choice model \citep{broder2012dynamic}, personalization \citep{cohen2020feature,ban2021personalized}, non-stationarity \citep{besbes2011minimax,chen2023learning}, ranking  \citep{gao2022joint} and discontinuity in the demand function \citep{den2020discontinuous}, among others.
Although the stream model is broadly applicable, in this work we aim to understand the pricing problem from a new perspective through the pool-based model.

We will soon see that every UDPM instance admits and optimal non-adaptive policy which is non-increasing; see \cref{prop:monotone}.
In revenue management, these policies are often referred to as {\em markdown} policies \citep{smith1998clearance,heching2002mark,caro2012clearance}.
A recent line of work studies markdown policies under the stream model, with unknown demand
\citep{chen2021multi,jia2021markdown,jia2022dynamic,salem2021algorithmic}.
Unlike our work, they view monotonicity as a {\bf constraint} rather than as a property derived from the model's intrinsic properties.

\subsubsection{Intertemporal Pricing.}
Intertemporal price discrimination has been studied since the seminal work of \citet{coase1972durability}, who postulates that a monopolist selling a durable good would have to sell the product at its marginal cost unless it commits to a pricing policy. 
The idea behind what is known as the {\em Coase conjecture} is that rational consumers know that the monopolist will reduce the price and therefore would wait until the price reaches the marginal cost. 
Henceforth, many works in economics focused on
characterizing optimal pricing policies, given that consumers strategically time their purchases (\citealt{coase1972durability}, \citealt{stokey1979intertemporal}, \citealt{conlisk1984cyclic}, and \citealt{sobel1991durable}).
This problem has also received significant attention in the revenue management community; see, e.g.,   \citealt{su2007intertemporal,aviv2008optimal,ahn2007pricing,correa2016contingent,wang2016intertemporal}.

However, the strategic behavior assumption is no longer reasonable in many applications.
For example, when the supply is finite or the good is perishable, consumers may not have the incentive to wait for a lower price \citep{correa2016contingent}. 
This motivates some work to incorporate myopic customers
\citep{besanko1990optimal,besbes2015intertemporal, lobel2020dynamic,caldentey2017intertemporal}.
These works assumed {\em continuous} monitoring of prices.
In contrast, in our model, customers monitor the price stochastically according to independent Poisson processes. 
\citealt{talluri2006theory} presented an LP formulation for computing an optimal policy for the pool-based model with $\lambda = \infty$ when given a limit on the number of price changes.

Other work has considered price-monitoring with finite frequency, usually by assuming a monitoring cost 
\citep{su2009value,aviv2008optimal,cachon2015price} and \cite{chen2018robust}. 
Empirically, \citep{moon2018randomized} estimated that consumers' opportunity costs for an online visit range from \$2 to \$25.
These works either focus on characterizing the equilibrium or assume a known model of the dynamics, whereas our work focuses on decision making under unknown demand.

Finally, there is a 
line of work 
which studies time-evolution of customer valuations.
\citet{garrett2011durable} and \citet{deb2014intertemporal} considered the impact on pricing strategies when customers' valuation evolves stochastically.
\citet{kakade2013optimal} and \citet{pavan2014dynamic} studied dynamic mechanism design where the buyer valuation changes over time, based on the private signals received.
\citet{chawla2016simple} investigated a pricing problem where a consumer's valuation evolves according to her previous experience as a martingale.
\citet{chen2019dynamic} studied a dynamic pricing problem in which the demand for a single product type in multiple local markets evolves linearly between two consecutive change points.
\citet{LemeSTW21} considered a dynamic pricing problem in which the buyer's valuation 
can change adversarially, but not too quickly.


\subsubsection{Robust Optimization in Pricing.}
In the non-adaptive setting, our problem (of optimizing the competitive ratio) boils down to a {\em robust optimization} problem, a framework commonly used in the revenue management literature; see, e.g., \citealt{ma2021dynamic, besbes2022beyond,wang2023power}.
Our non-adaptive setting may be most similar to the myopic customer setting in Section 4 of \citealt{caldentey2017intertemporal}.
They also assume that the seller only knows the support of the valuations, but not the actual valuations.
However, they assume that a customer makes a purchase {\em immediately} when the price is below the valuation, which is essentially a special case of our problem where the interaction rate approaches $\infty$.

In contrast, our work focuses on general $\lam$, which is substantially more challenging than the $\lam=\infty$ case.
In fact, with $\lam=\infty$, {\em any} policy that selects the prices in descending order is optimal. 
(Note, however, that this does not trivialize \citealt{caldentey2017intertemporal}'s results, as they consider customer arrivals.)
However, if $\lam<\infty$, the seller should make judicious decisions about how much time to spend at each price: 
If the price drops too quickly, a high-valuation customer may not have any interactions before the price reduces, incurring a revenue loss.
If the price decreases too slowly, the seller has less time to attract customers from the lower segments of the market.

\subsubsection{Partially Observable Reinforcement Learning.} 
Our problem can be reformulated as a {\em Markovian Decision Process} (MDP). 
In fact, we can \kehua\ the state by (i) the remaining customers in each valuation group and (ii) the remaining time.
A key challenge is that the seller only observes the total demand, but not the demand from each valuation group.
To address this, it is natural to introduce a prior distribution and reformulate this problem as a {\em partially observable MDP} (POMDP).
However, classical hardness results suggest that learning in POMDPs can be (both computationally and statistically) intractable even in simple settings \citep{krishnamurthy2016pac}.
Recent results for learning POMDPs are not applicable to our problem for multiple reasons. 
First, they are restricted to MDPs with special properties, such as {\em block MDPs} \citep{krishnamurthy2016pac,du2019provably} and {\em decodable MDPs} \citep{efroni2022provable}, but these 
do not hold in our setting.
Second, these methods typically depend on revisiting states, which is not feasible in our problem: state evolution is unidirectional, since the number of customers is non-increasing.

A related class of POMDP is the {\em restless bandits} problem \citep{whittle1988restless}, where the state of each arm evolves regardless of whether it is played.
This resembles the property of our problem that the remaining number of customers in every valuation group (``state'') may change even if the corresponding price (``arm'') is not selected.
However, in our problem, the transition probabilities depend on the arm selected, which is not true in restless bandits, rendering the existing results \citep{guha2010approximation,ortner2012regret} inapplicable.

%% file: OR/prelim.tex
{ We formally define our pool model and explain how it captures the stream model as a special case.

\subsection{The General Pool Model}

The seller is given a set $\mathcal{P} = \{p_1>\dots > p_k\}$ of prices to select from.
A pool of $n$ customers is present from the beginning.
Customer $i\in [n]$ has an {\em initial valuation} $v_i>0$.
After making $m$ purchases, her valuation becomes $\psi(m) v_i$ where the {\em discount function} $\psi: \mathbb Z_{\ge 0}\rar [0,1]$ is non-increasing.

Each customer interacts with the seller according to an \indep\ Poisson process over a finite time horizon $[0,1]$, (this does not lose generality since we can rescale $\lambda$ by the length of the horizon) with an identical rate $\lam>0$ for all customers. 
We call each event an {\em interaction} and $\lambda$ the {\em interaction rate}.
When an interaction occurs, the customer observes the price and makes a purchase if the price is less than or equal to her valuation.
The seller has unlimited inventory and knows both $\lambda$ and $\psi$.
We denote this instance by ${\rm Pool}(\lambda, \mathcal{P},{\bf v},\psi)$ where ${\bf v}=(v_i)_{i\in [n]}$. 

The seller aims to choose a sequence of prices to maximize the expected revenue.
The process of choosing prices is called a {\em policy}, which can be formally defined as a stochastic process $X=(X_t)_{t\in [0,1]}$ on ${\cal P}$.
We refer to a policy as {\em detail-free} or {\em detail-dependent}, depending on whether it relies on the knowledge of $\bf v$.
The {\em revenue} of a policy $X$ is $R_X:= \int_0^1 X_t\ dS_t$ where $S_t$ is the (random) sales up to time $t$. 

\subsection{Unifying the Pool and Stream Models}
To see the connection between the pool and stream models, let us rephrase the stream model \citep{gallego1994optimal}.
Customer interactions occur according to a Poisson process with a {\em base rate} $\lambda_{\rm base}$. 
In each interaction, the customer's valuation is drawn i.i.d., where we denote by $d(p)$ the \prb\ that the valuation is greater than or equal to $p\in \mathcal{P}$.
For any policy $(X_t)$, the demand rate at time $t$ is $d(X_t)\cdot \lambda_{\rm base}$.
We denote this instance by ${\rm GvR}(\lambda_{\rm base}, \mathcal{P},d)$.

We show that the stream model is a special case of the pool model.
\Sps\ the valuations are given by ${\bf v} = (v_i)_{i\in [n]}$. 
The {\em demand function} $d_{\bf v}:\cal P\rar \real$ is then given by $d_{\bf v}(p) = \frac 1n \sum_{i=1}^n \mathbbm{1}(v_i\ge p)$.
Next, we show that ${\rm Pool}(\lambda,{\cal P}, {\bf v},\psi\equiv 1)$ is equivalent to ${\rm GvR}(n\lambda, {\cal P}, d_{\bf v})$ in terms of expected revenue.


\begin{theorem}[Unifying Pool and Stream]\label{thm:unify_pool_and_stream} Let $X$ be any policy. For any $s,t\in [0,1]$, denote by $D_{s,t}$ the (random) demand in the time interval $[s,t]$.
Denote by $\ho{E}_{\rm GvR}$ and $\ho{E}_{\rm Pool}$ the expectation in instances ${\rm GvR}(\mathcal{P}, n\lambda, d_{\bf v})$ and ${\rm Pool}(\mathcal{P},\lam, {\bf v}, \psi\equiv 1)$ \resp. 
Then, \[\ho{E}_{\rm GvR}\lb[D_{s,t}\rb] = \ho{E}_{\rm Pool}\lb[D_{s,t}\rb].\]
\end{theorem}

To see this, \sps\ the price is $X_t$ in a small interval $[t,t+\delta] \sse [0,1]$.
Then, \begin{align}\label{eqn:122523}
\ho{E}_{\rm GvR} [D_{t,t+\delta}] \approx d(X_t)\cdot \lam_{\rm base}\delta =  d(X_t)\cdot n\lam\delta.
\end{align}
On the other hand, in the pool model, each customer monitors the price with \prb\ $\approx \lam\delta$.
Thus, 
\[\ho{E}_{\rm Pool}[D_{t,t+\delta}]\approx  \lambda \delta\sum_{i\in [n]} \mathbbm{1}(v_i\ge X_t)= \lam\delta \cdot d(X_t) n,\] which matches \cref{eqn:122523}.
We finally note that this equivalence holds in expectation, but the revenue distributions may differ.

\subsection{Our Focus: The Unit-Demand Pool Model (UDPM)}
In this work, we focus on the important special case where $\psi(m) = \mathbbm{1}(m=0)$, referred to as the {\em unit-demand pool model} (UDPM).
In this setting, each customer $i$'s valuation at any time is either $0$ or  the initial valuation $v_i$. 
Therefore, it loses no generality to assume that $v_i\in \cal P= \{p_j\}_{j\in [k]}$.
We denote by $n_j$ the number of customers with initial valuation $p_j$, which we refer to as {\em type-$j$} customers.
We denote by ${\rm UDPM}(\lam,\mathcal{P},(n_j)_{j\in [k]})$ the corresponding UDPM instance.

The problem is easy when the interaction rate is $\lam=\infty$. 
If the price is allowed to change arbitrarily many times, the problem is trivial, since any policy that selects all prices in $\cal P$ in decreasing order is optimal.
If the number of price changes is limited by a budget, the problem is non-trivial but \strfwd\ via a simple linear program (LP); see Section 5.5.1 in \citealt{talluri2006theory}.
Therefore, we focus on the \sce\ where $\lam <\infty$.

%% file: OR/known_demand.tex
We start with the \sce\ where the valuation distribution is known.
Unfortunately, unlike the stream model, where the optimal detail-dependent policy selects a fixed price, it is substantially more challenging to characterize the optimal policy in the pool model.
To see this, recall that in the MDP formulation, the ``state'' (i.e., number of remaining customers of each type) is not observable.
We therefore restrict ourselves to a simple yet compelling policy class where prices are determined in advance, regardless of the observations.

\begin{definition}[{\bf Non-adaptive Policy}] A policy $(X_s)_{s\in [0,1]}$ is {\em non-\adap} if for any $s\in [0,1]$, the \rv\ $X_s$ is a constant.
\end{definition}

Non-adaptive policies are widely applied in practice due to their simplicity
\citep{aviv2008optimal,yin2009optimal,chen2016real,caldentey2017intertemporal,chen2019efficacy,ma2021dynamic}.
We show how to efficiently find an optimal non-adaptive policy for any UDPM instance when the valuations are known. 

\subsection{Monotonicity of the Optimal Non-Adaptive Policy}
We first introduce a key structural result: 
we show that 
an optimal non-adaptive policy can be found within the class of markdown policies. 
Similar monotonicity results have been established in, for example, \citealt{gallego2008strategic,nasiry2011dynamic,adida2018markdown}.
This monotonicity property is obvious when $\lam=\infty$; 
Here, we establish this result for arbitrary $\lam$.

\begin{proposition}[Price Monotonicity]\label{prop:monotone}
There exists an optimal non-adaptive policy $(X_s)_{s\in [0,1]}$ such that $X_s \ge X_t$ a.s. whenever $0\le s < t\le 1$.
\end{proposition}


To see this, we apply a swapping argument. 
Suppose that the price is $p_L$ (``low'') in some time interval $[t-\eps,t]$ and increases to $p_H$ (``high'') in $[t, t+\eps]$.
We claim that the expected revenue does not decrease if we swap $p_H, p_L$.
To see this, consider a customer with valuation $v$.
If $v< p_L$, then the customer will not buy in $[t-\eps, t+\eps]$ regardless of the swap.
If $p_L\le v< p_H$, then the purchase \prb\ remains the same.
If $v\ge p_H$, then our revenue can only increase (or remain the same) if we offer $p_H$ earlier than $p_L$.

\brmk This monotonicity may not hold for optimal adaptive policies.
In fact, suppose that at some time, the market no longer contains low-valuation customers, yet high-valuation customers still remain.
Then, an optimal adaptive policy should increase the price. 
This suggests that adaptive policies are substantially more intricate than non-adaptive ones.\hfill \Halmos \ermk

We present a closed-form revenue function for any non-adaptive markdown policy, represented by a sequence ${\bf t} = (t_j)_{j \in [k]}$ with $\sum_{j=1}^k t_j = 1$. 
Here, $t_j$ represents the amount of time the policy selects $p_j$.

\begin{proposition}[Revenues of Non-adaptive Markdown Policies]\label{prop:exp_rev}
Consider a UDPM instance $\mathcal{I}={\rm UDPM} (\lambda,\{p_j\}_{j=1}^k,\{n_j\}_{j=1}^k)$.
For any non-adaptive markdown policy ${\bf t}= (t_j)_{j\in [k]}$, we have 
\begin{align}
\label{eqn:122523b}
{\rm Rev}({\bf t}, \mathcal{I}) := \ho{E}[R_X] = \sum_{\ell=1}^k n_\ell \sum_{j:\ell\le j} p_j e^{-\lambda \sum_{i = \ell}^{j - 1} t_i} \left(1-e^{-\lambda t_j}\right).
\end{align}
\end{proposition}

Each term in the inner summation of \cref{eqn:122523b} is the expected revenue from type-$\ell$ customers when we select $p_j$.
The term $e^{-\lambda \sum_{i = \ell}^{j-1} t_i}$ is the probability that a type-$\ell$ customer is still in the market (i.e., not made a purchase yet) when the policy switches to $p_j$.
The term $1-e^{-\lambda t_j}$ is the probability that a type-$\ell$ customer makes a purchase when the price is $p_j$, conditional on the event that she is still in the market. 

\subsection{Two Algorithms}
\label{sec:computing_opt}
By \cref{prop:monotone,prop:exp_rev}, finding an optimal non-adaptive policy reduces to 
\begin{align}\label{eq:max_non_adap}
& \max_{{\bf t} = (t_1,\dots,t_k)}  \quad \sum_{\ell=1}^k n_\ell \sum_{j:\ell\le j } p_j e^{-\lambda \sum_{i = \ell}^{j - 1} t_i} \left(1-e^{-\lambda t_j}\right) \notag\\
& \text{subject to}  \quad \sum_{j=1}^k t_j = 1,
\quad  t_j \ge 0, \quad \forall j \in [k].
\end{align}


Our first approach is based on dynamic programming (DP). We encode the {\em state}\footnote{
We note this is distinct from 
the ``state'' in the MDP formulation described earlier.} by (i) remaining time $t$, (ii) price $p_j$ under consideration, and (iii) the number $m$ of customers remaining in the market whose valuations are {\bf strictly} higher than $p_j$.
The remaining time can continuously vary in $[0,1]$.
Formally, we define the {\em value function} as 
\begin{equation} \label{eqn:value_func_base}
    \Phi(k,m,t) = p_k (m+n_k)\cdot (1-e^{-\lam t})
\end{equation} for any $m>0$ and $t\in [0,1]$.
Then for any $j\le k-1$, we recursively define 
\begin{equation} \label{eqn:value_func_recursive}
\Phi(j,m,t) = \max_{0 \le t' \le t} \lb\{p_j \left(m + n_j\right) \left(1 - e^{-\lambda t'}\right) + \Phi\lb(j +1, \left(m + n_{j}\right) e^{-\lambda t'}, t - t'\rb)\rb\}.
\end{equation}
Here, the function $\Phi(j,m,t)$ can be interpreted as the maximum expected revenue that a non-adaptive policy earns by marking down the price from $p_j$, with $m$ additional customers whose valuations are higher than $p_j$.
Specifically, we can view $t'$ as the amount of time that the policy selects $p_j$. 
The first term is the ``immediate reward'' (i.e., expected revenue at $p_j$) and the second is the maximum expected future revenue.

Our goal is to compute $\Phi(1,0,1)$. 
This can be done via a DP-based \alg\ using the recursion in \cref{eqn:value_func_base} and \cref{eqn:value_func_recursive}.
For efficient computation, we discretize the time horizon into small intervals of length $\eps$.
Since the function $1-e^{-\lam t}$ is $\lam$-Lipschitz, this leads to an $O(\lam \eps)$ discretization error in the value function.

Our second approach is based on continuous optimization. 
The global convergence of this approach is ensured by the concavity of the revenue function.

\begin{proposition}[Concavity of Revenue Function]
\label{prop:concave}
The function ${\rm Rev}({\bf t},{\cal I})$ is concave in $\bf t$ for any fixed UDPM instance ${\cal I}$.
\end{proposition}


{\fff In the case of either approach, we conclude the following theorem.}

\begin{theorem}[Computing Optimal Non-Adaptive Policy] \label{thm:optimal_non_adap}
For any $\eps>0$, given any UDPM instance ${\cal I}$ we can compute a non-adaptive policy ${\bf t}$ 
such that $|{\rm OPT}_{\rm NA}(\mathcal{I}) - {\rm Rev}({\bf t}, \mathcal{I})|= O(\lam \eps)$ in time polynomial in $k$ and $\frac 1\eps$.
\end{theorem}

So far, we have shown that the UDPM is tractable if the $(n_j)$'s are known.
In \cref{sec:non_adap,sec:adap}, we consider the case where the $(n_j)$'s are unknown.

%% file: OR/nonadap_V2.tex
One of the central challenges in dynamic pricing is demand uncertainty. 
From now on, we focus on detail-free policies, i.e., those that do not require knowledge of $(n_j)_{j\in [k]}$. 
We identify non-adaptive policies with robust performance, quantified by the competitive ratio. 

\bdefn[{\bf Competitive ratio}]
Let $\Pi$ be the class of all (possibly adaptive) policies and $\mathcal{F}$ be a family of UDPM instances.
The {\em competitive ratio} of a policy $X$ is
\[{\rm CR}(X, \mathcal{F}) := \inf_{\mathcal{I}\in \cal F} \frac{{\rm Rev}(X, \mathcal{I})}{{\rm OPT}(\mathcal{I})}\quad \text{where} \quad {\rm OPT}(\mathcal{I})= \sup_{X'\in \Pi} {\rm Rev}(X',\cI).\]
\edefn

When the family ${\cal F}$ of instances is clear, we will say a policy $X$ is {\em $c$-competitive} if ${\rm CR}(X, {\cal F}) \ge c$.
Note that the competitive ratio is defined for any policy $X$, regardless of its adaptivity, and measures the performance of $X$ relative to that of an optimal (possibly adaptive and detail-dependent) policy.

We first state a negative result.
Considering detail-free non-adaptive policies, we first show that the best-possible competitive ratio one could hope for is $1/k$.

\begin{theorem}[Upper Bound on the Competitive Ratio]\label{thm:non_adap_upper} Let ${\cal F}_k$ be the family of UDPM instances with $k$ prices. 
For any $\eps > 0, k \ge 1$ and detail-free non-adaptive policy ${\bf t}$, there is a UDPM instance $\cI_\eps\in \mathcal{F}_k$ s.t. \[{\rm Rev}({\bf t}, \mathcal{I}_\eps)\le \left(\frac 1k + \eps \right) {\rm OPT}(\mathcal{I}_\eps).\]
\end{theorem}
{\fff Essentially, this upper bound is proved by considering a Dirac distribution of valuations on a price that depends on $\cal P$.}
We defer the proof to Appendix \ref{apdx:UB}.  

{\fff \subsection{Two Simple (But Impractical) Policies} \label{sec:impractical_policies}}
The upper bound in \cref{thm:non_adap_upper} can be matched with a simple detail-free policy, called the {\em naive randomized} (NR) policy: Select a price uniformly at random and commit to it in the entire time horizon.

\begin{proposition} \label{prop:naive_randomized}
    The NR policy is $1/k$-competitive. 
\end{proposition}

{\fff However, this policy is highly impractical since it behaves in a pathological manner.}
\Sps\ all customer valuations are at the lowest price.
In this case, the NR policy chooses the optimal price w.p. $1/k$ but otherwise generates $0$ revenue (w.p. $1-1/k$).


{\fff A natural fix is to consider a deterministic analogue of the NR policy.  Consider the {\em naive deterministic} (ND) policy that allocates $1/k$ time to each price.}
Using Jensen's inequality, it
is not hard to show:

\begin{proposition} \label{prop:naive_deterministic}
The ND policy is $1/k$-competitive. 
\end{proposition}

Although this policy also meets the optimal $1/k$-bound, there are many cases {\fff where it performs poorly, since it does {\bf not} take into account the price set.
Consider a set ${\cal P}$ of $k\gg 2$ prices evenly spaced in the interval $[1,2]$. 
A simple policy that commits to the lowest price $p_k = 1$ is $1/2$-competitive, but the ND policy is only $1/k$-competitive (attained when all valuations are $1$).}

\subsection{A Robust Non-adaptive Policy}
In the rest of this section, we propose a policy with a simple closed-form formula that not only matches the upper bound on the competitive ratio, but also behaves more robustly compared to the NR and ND policy. 

\subsubsection{Max-Min Optimization.}
To proceed, we introduce an intuitive upper bound on the revenue of {\bf any} detail-dependent policy.
By Proposition \ref{prop:monotone}, we lose no generality by restricting ourselves to markdown policies.
For any instance $\mathcal{I} = (\lam, \mathcal{P}, (n_j)_{j\in [k]})$ and non-adaptive markdown policy ${\bf t} = (t_j)_{j\in [k]}$, consider  $f({\bf t}; n_1,\dots,n_k):= {\rm Rev}(X, \mathcal{I}) / {\rm OPT}(\mathcal{I}).$
Then, the problem of maximizing the CR is equivalent to
\begin{align*}
\text{(P0) } & \max_{t_1,\dots,t_k} \min_{n_1, \dots, n_k} \lb\{f(t_1,\dots,t_k; n_1,\dots,n_k): \sum_{j=1}^k n_j=n \rb\},\\
& \text{subject to} \quad \sum_{j=1}^k t_j = 1,\\
& \quad \quad \quad \quad \quad  t_j \ge 0, \quad \forall j \in [k].
\end{align*}

Unfortunately, it is challenging to directly solve (P0), since most results on max-min optimization rely on special structures such as concavity-convexity.
(To clarify, the function $f$ is not concave or convex. Please do not confuse this with the concavity of the revenue function in  \cref{prop:concave}, since the revenue function is only the numerator part of $f$.)
Furthermore, nonconvex-nonconcave min-max optimization is computationally hard \citep{daskalakis2021complexity}. 
An alternative idea is to find a closed-form formula for the ``min'' part given any $\bf t$, thus reducing (P0) to a single-level problem. 
However, finding a closed-form solution for ${\rm OPT}(\mathcal{I})$ is considerably formidable.

\subsubsection{Upper Bound on the Optimum}
We address the above challenge from a different perspective:
We will approximate the numerator and denominator of $f$, and transform (P0) into a tractable problem.
Then we argue that this transformation preserves the CR up to a factor of $k$.
We start by introducing an upper bound on ${\rm OPT}(\mathcal{I})$.

\begin{proposition}[Upper Bound on the Optimum]\label{prop:ub} 
For any (possibly adaptive or detail-dependent) policy $X$ and instance $\mathcal{I} = {\rm UDPM}(\lambda, \{p_i\}_{i=1}^k,\{n_i\}_{i=1}^k)$, we have
\[{\rm Rev}(X,\mathcal{I}) \le {\rm UB}(\mathcal{I}):= \sum_{i \in [k]} n_i p_i (1- e^{-\lambda}).\]
\end{proposition}

This \ineq\ is self-evident by observing that the expected revenue of a customer with a valuation $v$ is at most $v(1-e^{-\lambda})$.
For readers familiar with economics, ${\rm UB}(\cI)$ is {\fff the revenue achieved by} the {\em first-degree price discrimination}, modulo a factor of $1-e^{-\lam}$.

\brmk This upper bound becomes tighter as $\lam$ increases.
To see this, observe that when $\lam = \infty$, we can achieve (perfect) first-degree price discrimination by moving from the highest to the lowest price, hence the identity holds.
On the other extreme, UB can be very loose when $\lam$ is close to $0$, essentially because  ${\rm UB}(\mathcal{I})$ is achieved by a {\em personalized} policy while ${\rm OPT}(\mathcal{I})$ is defined over {\em non-personalized} policies.
Next, we construct a non-adaptive policy whose revenue is $\Omg\left(\frac 1k \right) \cdot \mathrm{UB}(\mathcal{I})$, which implies that UB approximates the optimum within a factor of $k$.\hfill\Halmos\ermk 
{\fff Next, we derive a robust policy based on this upper bound.}

\subsubsection{A Robust Non-adaptive Policy}
The above upper bound motivates us to consider a simpler function $\tilde f(t_1,\dots, t_k;n_1,\dots,n_k) := {\rm Rev}(\mathcal{I}) / {\rm UB}(\mathcal{I})$ as a surrogate for the original objective $f$. 
Unfortunately, $\tilde f$ is the ratio of two non-linear functions, and so the corresponding max-min problem is still challenging to solve.

We address this by {\em linearizing} both the numerator and denominator, thereby converting the max-min optimization to an LP.
The above sequence of reductions loses only a factor of $k$ (see Appendix \ref{apdx:non_adap}). 
We then derive the following robust non-adaptive policy by finding a closed-form solution to the resulting LP.
{ \bdefn[{\bf Robust Policy}]\label{def:non_adap}
Given any price set $\mathcal{P}=\{p_1,\dots,p_k\}$, the {\em robust policy} ${\bf t}^\star_{\mathcal{P}} =(t^\star_j)_{j\in [k]}$ is
\begin{align} \label{eqn:lp_characterization}
t_k^\star = \frac{1}{k - \sum_{j=1}^{k-1} \frac{p_{j+1}}{p_{j}}} \quad \quad \text{and}\quad \quad  t_{j}^\star = \left(1 - \frac{p_{j+1}}{p_{j}} \right) t_k^\star, \quad \forall j=1,\dots, k-1.
\end{align}
\edefn

For example, for $\mathcal{P} = \left\{1,2^{-1},\dots,2^{-(k-1)}\right\}$, we have $t_k^\star = 2/(k + 1)$ and $t^\star_j = 1/(k+1)$ for $j=1,\dots,k-1$, i.e., the policy spends $1/(k+1)$ time on each of the first $(k-1)$ prices and $2/(k+1)$ time on the last. 
{\fff This robust policy also matches the upper bound on the CR.}

\begin{theorem}[Competitive Non-adaptive Policy]\label{thm:non-adap}
Denote by $\mathcal{F}_{\mathcal{P}}$ the family of UDPM instances with price set $\mathcal{P}$.
Then, 
\begin{align}\label{eqn:122723}
{\rm CR}({\bf t}^\star_{\mathcal{P}}, \mathcal{F}_{\mathcal{P}})\ge \frac{1}{k - \sum_{j=1}^{k-1} \frac {p_{j+1}}{p_j}} \ge \frac 1k.
\end{align}
\end{theorem}

{\fff 
\brmk The second \ineq\ can be very loose. 
Consider again the set ${\cal P}$ of $k$ prices evenly spaced in $[1,2]$.
The ND policy has a CR of $\frac{1 - e^{-\lam/k}}{1 - e^{-\lam}}\approx \frac 1k$  when $\lambda$ is small.
In contrast, our robust policy has a CR of \[\frac 1{k - \sum_{j=1}^{k-1} \frac {p_{j+1}}{p_j}} = \frac 1 {k- \sum_{j=1}^{k-1}\lb(\frac{2 - (j+1)/k}{2 - j/k}\rb)} = \frac 1 {k - \sum_{j=1}^{k-1} \left(1 - \frac{1}{k - j}\right)}\approx \frac 1{1+\log k}\]
for $\cal P$ when $k$ is large, which is much better than the ND policy.\hfill\Halmos
\ermk}

As a by-product,   \cref{thm:non-adap} implies a bound on the adaptivity gap.
Let $\Pi_{\rm NA}$ be the class of non-adaptive detail-free policies and $\Pi$ the class of (any) detail-free policies.
Let $\mathcal{F}_k$ be the set of UDPM instances with $k$ prices. 
For any instance $\cal I$, denote ${\rm OPT}_{\rm NA}(\mathcal{I}) = \sup_{X\in \Pi_{\rm NA}} {\rm Rev}(X, \mathcal{I})$.
Then,
\[{\rm Rev}({\bf t}^\star_{\cal P}, \cal I)\le {\rm OPT}_{\rm NA}(\cal I) \quad \text{and}\quad   {\rm UB}(\cal I)\ge {\rm OPT}(\mathcal{I}),\]
and therefore Theorem \ref{thm:non-adap} immediately implies the following.

\bcoro[Adaptivity Gap] For any $k\ge 1$ and $\mathcal{I} \in \mathcal{F}_k$, we have
\[{{\rm OPT}(\mathcal{I})}\le k\cdot {\rm OPT}_{\rm NA}(\mathcal{I}).\]
\ecoro

{\fff Next, we extend our results to handle continuous price space.} 

\begin{wrapfigure}{R}{10cm}
\centering
\includegraphics[width=10cm, height = 4.5cm]
{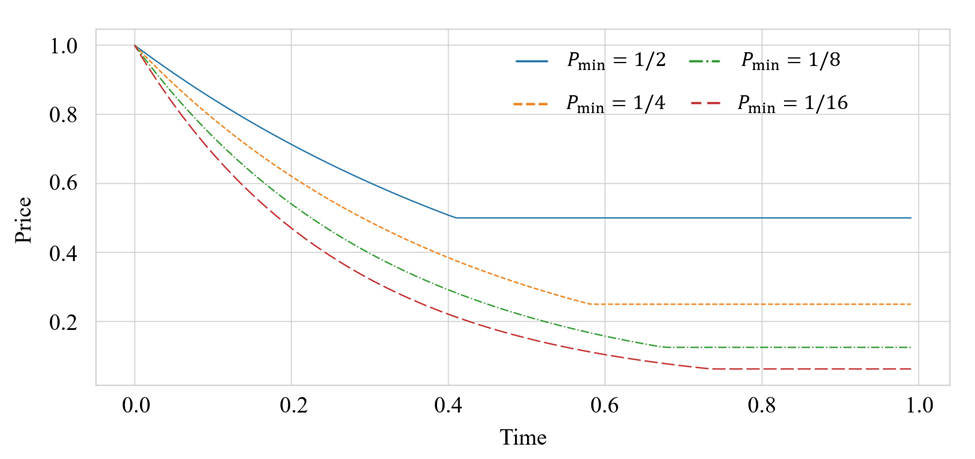}
\caption{Robust non-adaptive policies for a continuous price space $[p_{\min},1]$.}
\label{fig:price_time}
\end{wrapfigure}

\subsection{Extending to Continuous Price Intervals} \Sps\ $\cal P=[a, b]$. For any $\eps > 0$, we define the {\em $\eps$-geometric grid} as $\overline {\cal P}_\eps = \{a_\ell :=(1 + \eps)^\ell \mid \ell = 1,\dots, \lceil \log_{1+\eps}(b/a)\rceil\}.$
We now analyze the CR of the robust policy (given in \cref{def:non_adap}) on $\overline {\cal P}_\eps$. 

By restricting 
to the geometric grid, we lose a (multiplicative) factor of $(1 + \eps)$ in the revenue. Thus, 

\begin{align}\label{eqn:122923}
\frac{{\rm Rev}({\bf t}^\star_{\overline{\mathcal{P}}_\eps}, \mathcal{I})}{{\rm UB}(\mathcal{I})} 
&\ge \frac 1{(1 + \epsilon)\lb(k - \sum_{i=1}^{k-1} \frac{p_{i+1}}{p_{i}}\rb)}\notag\\
&= \frac 1{(1+\eps)(k-\frac {k-1}{1+\eps})}\notag\\
&= \frac 1{1+\eps k}\notag\\
&= \frac 1{1+ \eps \log_{1+\eps}\frac {p_{\max}}{p_{\min}}}\notag\\
&\ge \lim_{\eps\rar 0^+}\frac 1{1+\ln \frac {p_{\max}}{p_{\min}}},
\end{align} where the last step is because $\eps/\ln (1+\eps)$ increases in $\eps$.
We summarize the above as follows.

\bdefn[{\bf Robust Policy for Continuous Price Space}]
Given the price interval $[p_{\rm min},p_{\rm max}]$, we define the {\em robust non-adaptive policy} $X=(X_t)$ as
\begin{align}\label{eqn:031424}
X_t = 
\begin{cases}
     p_{\max} \cdot e^{-t \lb(1 + \ln  \frac{p_{\max}}{p_{\min}}\rb)}, & \text{if  } 0 \le t \le 1 - \frac 1{1 + \ln  \frac{p_{\max}}{p_{\min}}}\\
    p_{\min}, & \text{if  }  1 - \frac{1}{1 + \ln  \frac{p_{\max}}{p_{\min}}} < t \le 1.
\end{cases}
\end{align}
\edefn

\begin{theorem}[Competitive Ratio for Price Intervals]\label{thm:conti_P}
For any UDPM \ins\ $\cal I$ with price set $\cal P = [p_{\rm min},p_{\rm max}]$, the robust policy $X$ \sats\
\[\frac{{\rm Rev}(X, \mathcal{I})}{{\rm UB}(\mathcal{I})} \ge \frac 1{1 + \ln\frac{p_{\max}}{p_{\min}}}.\]
\end{theorem}
}

{ 
We visualize the robust non-adaptive policies in \cref{fig:price_time}.
The price curves become steeper as $p_{\min}$ decreases and when $t$ approaches 1, the price reaches $p_{\min}$ and remains constant at $p_{\min}$ until the end of the horizon.
This is consistent with our intuition. 
In fact, when $p_{\min}$ is small, the ``opportunity cost'' to choose the lowest price is relatively small. 

}

%% file: OR/Learning-New.tex
{
Now we consider adaptive policies in the presence of unknown demand.  
Specifically, the seller only knows that the underlying \ins\ $\cal I$ comes from the family ${\cal F}(\lam,{\cal P},n)$ of instances with interaction rate $\lam$, price set $\cal P$ and a total number $n$ of customers.
In \parti, the seller does not know $(n_j)_{j\in [k]}$. 
{\fff 
As discussed in \cref{sec:known_demand}, the structure of the optimal adaptive policy is substantially more delicate (in \parti, it may not be markdown). We therefore compare} our policy against an optimal {\em non-adaptive} policy. 
By abuse of notation, we denote by $\opt(\cI)$ instead of ${\rm OPT}_{\rm NA} ({\cal I})$.

\begin{definition}[{\bf Regret}]
For any UDPM \ins\ ${\cal I} = {\rm UDPM}(\lambda, {\cal P}, (n_j)_{j \in[k]})$, we define the {\em regret} of a policy $X$ as 
$\textrm{Regret}(X, {\cal I}) := {\rm OPT}({\cal I}) - \E[R_X]$ where $R_X$ is the revenue of $X$.
\end{definition}

\subsection{The Learn-then-Earn Policy}
We present a {\em learn-then-earn} (LTE) policy (which is adaptive and detail-free) with sublinear regret in {\bf both} $n$ and $k$.
The policy operates in two phases: It first learns the $(n_j)$'s by exploring each price, then commits to an optimal non-adaptive policy for the learned instance.
More precisely, 
\bitem 
\item In the learning phase, we explore each price $p_j$ for a fixed interval of length $s_j$ in decreasing order.
Then, we construct estimates $(\widehat n_j)_{j \in [k]}$ of $(n_j)_{j \in [k]}$ using the realized demands $D_j$ based on a novel debiasing approach (see \cref{sec:debias}).
\item  In the earning phase, we compute a (nearly) optimal non-adaptive policy $\widehat{\bf t}$ for the estimated instance $\widehat {\cal I} = (\lambda, {\cal P}, \{\widehat n_j\}_{j \in [k]})$, which can be done efficiently per our discussion in \cref{sec:computing_opt}.
We note that since we used $\sum_{j=1}^k s_j < 1$ time to learn the instance, we will shrink every interval in $\widehat{\bf t}$ by a $(1-\sum_{j=1}^k s_j)$-factor. 
\eitem

{\fff We formally state the policy in \cref{alg:learn_then_earn}. 
As the main result of this section, we show that this policy achieves sublinear regret in both $n,k$.}

\begin{theorem}[Adaptive Policy with Sublinear Regret]
\label{thm:regret_bound}
For any $\lam, {\cal P}, n$, there exist parameters for \cref{alg:learn_then_earn} such that for any
${\cal I}\in {\cal F}(\lam, {\cal P}, n)$, we have \[{\rm Regret}(X^{\rm LTE}, {\cal I}) = \tilde O(p_1 (kn)^{2/3}).\]
\end{theorem}

\brmk 
Our regret bound is higher than the minimax optimal $\tilde \Theta(\sqrt{kn})$ regret \citep{auer2002finite} for the stream model (which is equivalent to $k$-armed bandits). 
Intuitively, this is because in the stream model, the effect of exploration is {\bf local}: The seller's decision in any time interval only affects the customer interactions that arrive in the interval.
In contrast, in our model, a price decision has a {\bf global} impact: A poor price choice potentially affects $\Omg(n)$ customers, so it is reasonable to expect a higher regret.
\hfill \Halmos \ermk

Although this is similar to the {\em Explore-Then-Commit} (ETC) policy in MAB, there are several new challenges in the analysis, as we will explain in the rest of this section.

\begin{algorithm}
\caption{Learn-Then-Earn (LTE) Policy \label{alg:learn_then_earn}}
\DontPrintSemicolon
\SetKwComment{Comment}{//}{}
\KwData{Partial Instance  $(\lambda, {\cal P}, n)$, \ Exploration times $(s_1, s_2, \ldots, s_{k})$}
\KwResult{Policy $X^{\rm LTE}$}

\Comment*[l]{Learning phase}
\For{$j = 1,2,\ldots, k$}{
Select $p_j$ for time $s_j$ and observe sales $D_j$
}

\For{$j=1,2,\ldots,k$}
{$\widehat n_j \gets \frac{D_j}{q(s_j)} - \sum_{i\le j-1} (\widehat n_i - D_i)$ where $q(x) := 1-\exp(-\lambda x)$
}

\Comment*[l]{Earning Phase}
$\widehat {\cal I} \gets
{\rm UDPM}\lb(\lambda, {\cal P}, \{\widehat n_j\}_{j \in [k]}\rb)$

Compute an optimal non-adaptive policy $\widehat{\bf t} = (\widehat t_j)_{j \in [k]}$ for $\widehat \cI$ using \cref{thm:optimal_non_adap}.

$s_{\rm total} \gets \sum_{j=1}^{k} s_j$

\For{$j=1,2,\ldots,k$}{
Select price $p_j$ in the time interval $[(1-s_{\rm total})\widehat t_j + s_{\rm total}, (1-s_{\rm total})\widehat t_{j+1} + s_{\rm total}]$}
\end{algorithm}

\subsection{Key Component: Unbiased Estimator for the Valuations} \label{sec:debias}
As is standard in MAB, the key is to construct \emph{unbiased} estimates of $(n_j)$'s.
This is easy for $n_1$. Recall that in Algorithm \ref{alg:learn_then_earn}, we selected $p_1$ in the time interval $[0,s_1]$ and observed a (random) demand of $D_1$. 
Note that $D_1 \sim {\rm Bin}(n_1, q(s_1))$ where $q(x) := 1-\exp(-\lambda x)$, so $D_1 / q(s_1)$ is an unbiased estimate of $n_1$.

However, the situation becomes more complicated when we explore lower prices due to {\bf confounding}.
In fact, when we select $p_2$, there may still be type-$1$ customers who have not yet exited the market. 
Therefore, both type-$1$ and type-$2$ customers contribute to the demand $D_2$ observed at the price $p_2$.

We derive a novel unbiased estimator to overcome this. 
Starting with $\widehat n_1 = D_1 / q(s_1)$, for each $j =2,3,\dots,k$, we recursively define 
\begin{equation} \label{eqn:estimates}
\widehat n_j = \frac{D_j}{q(s_j)} - \sum_{i\le j-1}\lb(\widehat n_i - D_i\rb),
\end{equation}
The first part is similar to the naive estimator for $\widehat n_1$, while the second part aims to ``deconfound'' the signals from  customers with higher valuations.
We show that this estimator is unbiased.

\begin{lemma}[Unbiasedness] \label{lem:unbiased_estimates}
For any UDPM \ins\ ${\cal I}$ and policy \pmt s $s_1, s_2, \ldots, s_{k}$, we have $\ho{E} [\widehat n_j] = n_j$ for any $j=1,\dots,k$.
\end{lemma}
}

\proof {Proof Sketch.}
For any $j\in [k]$, we have $D_j \sim {\rm Bin}(\sum_{i \leq j} n_i - \sum_{i\le j-1} D_i,\  q(s_j))$, so
\[\ho{E}[\widehat n_j \mid D_1,\ldots ,D_{j-1}] = \sum_{i \leq j} n_j - \sum_{i \leq j-1} D_i - \sum_{i \leq j-1} (\widehat n_i - D_i) = n_j +  \sum_{i \leq j-1} \lb(n_i - \widehat n_i\rb),\]
Therefore, by the law of total \prb, $\ho{E} [\widehat n_j] = n_j  + \ho{E}[\sum_{i\le j-1} (n_i - \widehat n_i)]$. 
The second term is $0$ by induction \hypo.\hfill \Halmos

{ 
\subsection{Warm-up: The Two-Price Setting} \label{sec:two_price_case}
We illustrate the main ideas by considering $k=2$.
In this setting,  \cref{alg:learn_then_earn} chooses $p_1$ throughout the learning phase and observes demand $D_1$.
As discussed in \cref{sec:debias}, $\widehat n_1 = D_1/ q(s)$ and $\widehat n_2 = n - \widehat n_1$ are unbiased estimates of $n_1$ and $n_2$. 
Using these, we compute an optimal non-adaptive policy $\widehat {\bf t} = (\widehat t_1, \widehat t_2)$ for the estimated instance $\widehat {\cal I} = (\lambda, {\cal P}, (\widehat n_1, \widehat n_2))$. 
Finally, since the remaining time is $1-s$, we implement the policy $(1-s)\bf \widehat t$, in which each price $p_i$ is selected for $(1-s)\widehat t_i$ time.

To analyze the regret, we first decompose it into two parts. 

\bdefn[{\bf Regret Due to Learning and Estimation}]
The {\em remaining} instance is defined as ${\cal I}_{\rm rem} = (\lambda, {\cal P}, \{N_1, N_2\})$ where $N_j$ ($j=1,2$) is the (random) number of type-$j$  customers remaining after the learning phase.
Let ${\bf t^\star}$ be an optimal non-adaptive policy for $\cI$.
We define \[\eta_1 = \lb|\E\lb[{\rm Rev}( (1-s)\widehat {\bf t},\ {\cal I}_{\rm rem})\rb] - \E\lb[{\rm Rev}\lb(\widehat {\bf t}, \cI\rb)\rb]\rb|\] 
and
\[\eta_2 = \lb|\E\lb[{\rm Rev}\lb(\widehat {\bf t}, \cI\rb)\rb] - \opt(\cI)\rb| = \lb|\E\lb[{\rm Rev}\lb(\widehat {\bf t}, \cI\rb)\rb] - \E \lb[{\rm Rev}\lb({\bf t}^\star, \cI\rb) \rb]\rb|.\]
\edefn 

{\fff We can view $\eta_2$ as the regret due to estimation error. 
The interpretation of $\eta_1$ may be less obvious. 
It is the regret due to learning - the difference in the revenue if we apply ${\bf \wh t}$ on $\cI_{\rm rem}$ versus on $\cI$. 
Using the above, we can isolate these two sources of regret:}

\begin{lemma}[Regret Decomposition]\label{lem:regret_decomp_two_price}
For any UDPM instance $\cI$, we have ${\rm Regret}(X^{\rm LTE}, \cI) \le \eta_1 + \eta_2.$
\end{lemma}

Unsurprisingly, $\eta_1$ is at most linear in $s$. 

\begin{lemma}[Bounding $\eta_1$] \label{lem:eta1_bound}
It holds that $\eta_1 = O(\lambda n p_1 s)$.
\end{lemma}
\proof {Proof Sketch.}
The number of customers who made a purchase during the learning phase is $O(n\lam s)$.
Therefore, the number of customers in $\cI$ and $\cI_{\rm rem}$ differ by $O(n\lam s)$. 
Next, we show that $\bf\widehat t$ and $(1-s)\bf \widehat t$ differ by $O(n\lam s)$ in their expected revenue. 
In fact, in the $j$-th time interval of the earning phase earned, the revenues of these two policies differ by at most $| e^{-\lam \widehat t_j} - e^{-(1-s)\lam \widehat t_j}| = e^{-\lam \widehat t_j} (e^{\lam s\widehat t_j} - 1) = O(\lam s\widehat t_j).$\hfill \Halmos

We next bound $\eta_2$. Note that $D_1$ is an i.i.d. sum of $n_1$ Bernoulli variables with mean $q(s)$, we can derive high-\prb\ bounds on $|n_1 - \widehat n_1|$ using concentration bounds. 
A naive application of Hoeffding's \ineq\ implies that $\eta_2 = O(\frac{\sqrt{n \log n}}{\lambda s})$, which leads to a subpar regret bound of $\tilde O(n^{3/4})$.
However, this bound can be substantially improved. 
In fact, Hoeffding's \ineq\ neglects the fact that $q(s)=o(1)$ as $n\rar \infty$.
To utilize this, we apply Bernstein's inequality and derive the following stronger bound:

\begin{lemma}[Bounding $\eta_2$] \label{lem:eta2_bound}
We have $\eta_2 = O\lb(p_1\sqrt{\frac {n \log n}{\lambda s}}\rb)$.
\end{lemma}

Combining  \cref{lem:regret_decomp_two_price}, \cref{lem:eta1_bound} and \cref{lem:eta2_bound}, we obtain
\[{\rm Regret}(X^{\rm LTE}, {\cal I}) \leq O\left(\lambda n p_1 s +  p_1 \sqrt{\frac {n \log n}{\lambda s}}\right).
\]
Taking $s=n^{-1/3} / \lambda$, we immediately obtain the $\tilde O(n^{2/3})$ regret bound.
}

{ 
\subsection{Extending to $k> 2$}
The analysis for general $k$ is more involved, since we need to carefully analyze how the estimation errors accumulate.
Rather than bounding the estimation errors $\widehat n_j - n_j$ individually, we prove a novel concentration inequality for the total estimation error $\sum_{j \in [k]}(\widehat n_j - n_j)$.  
The first challenge is dependence: The estimator $\widehat n_j$ depends on $\widehat n_1,\dots, \widehat n_{j-1}$. 
Additionally, we will see that the process $Y_j = \sum_{i \leq j} (\widehat n_i - n_i)$ does not form a martingale, so standard tools such as Azuma-Hoeffding's inequality do not apply either. 
We address this by showing that the process admits a {\em self-neutralizing} property that allows us to apply the Cram{\'e}r-Chernoff method to obtain a concentration bound.

As in the $k=2$ setting, we start by decomposing the regret. 
Let $\widehat {\cal I}$ be the estimated instance given in  \cref{alg:learn_then_earn}. 
Analogously to the $k=2$ setting, we define the {\em remaining instance} as ${\cal I}_{\rm rem}= {\rm UDPM}(\lambda, {\cal P}, (N_j)_{j \in [k]})$ where $N_j$ is the (random) number of type-$j$ customers remaining after the learning phase.  
Then, the regret can be decomposed as $\eta_1 + \eta_2$ where \[\eta_1 = \lb|\E \lb[{\rm Rev}\lb( \lb(1-\sum_{j=1}^k s_j\rb)\widehat {\bf t},\  {\cal I}_{\rm rem}\rb)\rb] - \E\lb[{\rm Rev}\lb (\widehat {\bf t}, \cI\rb)\rb]\rb| \quad\text{and}\quad \eta_2 = \lb|\E \lb[{\rm Rev}\lb(\widehat {\bf t}, \cI\rb)\rb] - \opt(\cI)\rb|.\]  
As in the $k=2$ setting, it is \strfwd\ to show that $\eta_1$ is linear in the length of the learning phase. 

\begin{lemma}[Bounding $\eta_1$] \label{lem:k_price_eta1_bound}
We have $\eta_1 = O(\lambda n p_1 \sum_{j=1}^k s_j)$.
\end{lemma}
The main {\bf challenge} is to bound $\eta_2$. 
We need to ensure that the estimation errors accumulate slowly.
We devote the remainder of this section to showing a bound on $\eta_2$ that is sublinear in {\bf both} $n$ and $k$.

\subsubsection{Self-Neutralizing Process and Conditional Subgaussianity}
Our analysis of $\eta_2$ is based on two key properties of the error process. 
Consider a \stoch\ process ${\bf \Delta} = (\Delta_j)_{j \in [k]}$ (which will be chosen as $\Delta_j = \widehat n_j - n_j$) adapted to the filtration $({\cal F}_j)_{j \in [k]}$ where ${\cal F}_j= \sigma(D_1,\dots,D_{j-1})$.
The first property states that this process has a tendency to move towards the origin.

\begin{definition}[{\bf Self-Neutralizing Property}] \label{def:self_neutralizing_process}
The process ${\bf \Delta}$ is {\em self-neutralizing} (SN) if for all $j \in [k]$ we have
$\E[\Delta_j \mid {\cal F}_{j-1}] = - \sum_{i \leq j-1} \Delta_i.$
\end{definition}

To control the deviation, we also need to bound the moment generating function (mgf).

\begin{definition}[{\bf Conditional Sub-gaussianity}] \label{def:conditional_subgaussian}
A process ${\bf \Delta}$ is {\em conditionally sub-gaussian} if there exist real numbers $(c_j)_{j \in [k]}$ such that for all $j \in [k]$ and $\gamma > 0$,  
\begin{align}\label{eqn:022824}
\E\lb[e^{\gamma \Delta_j} \mid {\cal F}_{j-1}\rb] \le \exp \left( \gamma \E[\Delta_j \mid {\cal F}_{j-1}] + \frac{\gamma^2 c_j^2}2 \right).
\end{align}
\end{definition}

In \parti, if the process is also self-neutralizing, then \cref{eqn:022824} is equivalent to 
\[\E\lb[e^{\gamma \sum_{i=1}^j\Delta_i} \mid {\cal F}_{j-1}\rb] \le \exp \left(\frac{\gamma^2 c_j^2}2 \right).\]
As we will explain in \cref{subsec:estimation_error}, the estimation error can be viewed as a weighted sum of a conditionally sub-gaussian SN process. This motivates the following concentration bound for such processes. 

\begin{theorem}[MGF Bound] \label{thm:mgf_bound}
Let ${\bf \Delta}= (\Delta_j)_{j \in [k]}$ be a self-neutralizing and conditionally sub-gaussian process with constants $(c_j)_{j \in [k]}$.
Then, for any $\gamma > 0$ and non-increasing sequence ${\bf r}=(r_j)_{j\in [k]}$, we have
\begin{align}\label{eqn:022824b}
\E\lb[e^{\gamma \lan {\bf r},{\bf \Delta}\ran}\rb] \leq \exp\left(\frac {\gamma^2}2 \sum_{j = 1}^k (r_j c_j)^2 \right).
\end{align}
\end{theorem}

\brmk The above bound resembles the Azuma-Hoeffding inequality for subgaussian \mtg\ difference sequences; see e.g., Theorem 2.19 of \cite{wainwright2019high}. 
However, we reiterate that $\Delta_j$ is not a super/sub-\mtg, so the Azuma-Hoeffding inequality does not apply. 
Instead, we prove the above by a careful induction in Appendix \ref{apdx:mgf_bd}.
We also emphasize that the monotonicity in $\bf r$ is necessary, since our induction step considers a new non-increasing sequence ${\bf r}'=(r_j')$ where $r_j' = r_j - r_k \ge 0$.\hfill \Halmos\ermk 

\subsubsection{Bounding the Deviation of Estimation Error}\label{subsec:estimation_error}
We next use \cref{thm:mgf_bound}  to control the estimation error in the expected revenue for a fixed non-adaptive policy.
To see this connection, we write the revenue function as the inner product of a non-increasing vector and a conditionally sub-gaussian SN process as follows.
Observe that for any UDPM instance $\cI$ and non-adaptive policy ${\bf t}$, we have
\[{\rm Rev}({\bf t}, \cI) = \sum_{j=1}^k r_j({\bf t}) \cdot n_j,\] 
where $ r_j({\bf t}) $ is the expected revenue from type-$j$ customers, formally given by \[r_j({\bf t}) = \sum_{i=j}^k e^{-\lam \sum_{\ell =j}^{i-1} t_\ell} \lb(1-e^{\lam t_i}\rb) p_i.\]
Thus, we can write the estimation error as
\begin{equation}\label{eqn:022624}
{\rm Rev}\lb({\bf t},\ \widehat \cI\rb) - {\rm Rev}({\bf t},\ \cI) = \sum_{j=1}^k r_j({\bf t})(\widehat n_j - n_j) = \lan {\bf r,\ \widehat n - n}\ran.
\end{equation}
which matches the exponent in \cref{eqn:022824b}.

This enables us to apply the mgf bound in \cref{thm:mgf_bound} on  \cref{eqn:022624} and hence control the estimation error of ${\rm Rev}({\bf t},\ \cI)$.
To this end, we verify that the process $\{\widehat n_j - n_j\}_{j\in [k]}$ is self-neutralizing and conditional sub-gaussian with $c_j^2 = 2\sum_{i \leq j} n_i / q(s_j)$; see \cref{lem:errors_self_neutralizing} and \cref{lem:errors_conditionally_subgaussian} in Appendix \ref{apdx:error_process}.
Moreover, ${\bf r} = (r_j({\bf t}))_{j\in [k]}$ is non-increasing for any fixed $\bf t$.
Thus, by \cref{thm:mgf_bound} and Markov's \ineq,
\begin{align}\label{eqn:022824f}
\ho{P}\lb[\lan {\bf r,\ \widehat n - n}\ran> \tau \rb] 
\le e^{-\gamma \tau}\cdot \ho{E} \lb[e^{\gamma \lan {\bf r,\ \widehat n - n}\ran}\rb]
\le  e^{-\gamma \tau}\cdot \exp\lb(-\frac{\tau^2}{4 \sum_{j=1}^k r_j^2 c_j^2 } \rb).
\end{align}
Taking $\gamma = \frac \tau {2\sum_{j=1} c_j^2 r_j^2}$ and recalling that $c_j^2 = 2\sum_{i \leq j} n_i / q(s_j)$, we have
\begin{align*}
\eqref{eqn:022824f}
= \exp\lb(-\frac{\tau^2}{4 \sum_{j=1}^k r_j^2\sum_{i\leq j} \frac{n_i}{q(s_j)} } \rb) 
\le \exp\lb( -\frac{\tau^2}{4 r_1^2n \sum_{j=1}^k \frac{1}{q(s_j)} } \rb),
\end{align*}
where the inequality follows since $\sum_{i \leq j} n_i \le n$.
By \sym, we can obtain a similar upper bound on $\ho{P}\lb[\lan {\bf r,\  \widehat n - n}\ran < -\tau \rb]$.
The above can be summarized as follows.

\bprop[Concentration Bound for the Error Process]
\label{prop:pointwise}
Let $\bf t$ be any non-adaptive policy and write ${\bf r}=(r_j({\bf t}))_{j\in [k]}$. 
Then, for any $\tau>0$, 
\[ \ho{P} \lb[ \lb|{\rm Rev}\lb({\bf t},\ \widehat \cI\rb) - {\rm Rev}({\bf t},\ \cI)\rb| > \tau \rb]  = \ho{P} \lb[ \lb| \lan {\bf r,\ \widehat n - n}\ran \rb| > \tau \rb] 
\le 2 \exp\lb( -\frac{\tau^2}{4 r_1^2n \sum_{j=1}^k \frac{1}{q(s_j)} } \rb).\]
\eprop

Next, we explain how to use the above to bound the regret in the earning phase.

\subsubsection{From Pointwise to Uniform Concentration}
So far, we have bounded the estimation error for each non-adaptive policy {\em individually} (``pointwise''). 
However, this is not sufficient to conclude the analysis. 
To see this, let us recall the regret analysis of the LTE policy in the stream model (which is equivalent to MAB). 
In the learning phase, we obtain an estimate $\hat \mu_a$ of the mean reward $\mu_a$ for each arm $a\in [k]$. 
In the earning phase, by committing to the empirically optimal arm, we can bound the regret using $\max_{a\in [k]} |\mu_a - \widehat \mu_a|$, which is controlled by a standard combination of union bound and concentration bounds.

The above suggests that the key is to control is the maximum estimation error of a set of ``arms'' (policies, in our context). 
Unfortunately, in our pool model, there are infinitely many ``arms''.
One is tempted to discretize the policy space (which is a $(k-1)$-dimensional simplex), but this involves $m=e^{\Omg(\sqrt k)}$ terms, which leads to an extra $\log m = \Omg(\sqrt k)$ regret term.
Another idea is to extend the above analysis to random sequences of $r_j$'s that depends on {\bf all} $\widehat n_j$'s.
However, \cref{thm:mgf_bound} requires that ${\bf r}$ be independent of the error sequence ${\bf \Delta}$.
We address this using the following structure of the policy space.



\begin{lemma}[Basis of the Policy Space] 
\label{lem:convex_comb_argument} 
Denote by $(e_j)_{j\in [k]}$ the natural basis for $\real^k$.
For each $j\in \{0\} \cup [k]$, define $\phi_j = \sum_{i=1}^j e_i$. 
Then, any non-increasing sequence ${\bf r}\in [0,1]^k$ is a convex combination of $\{\phi_j\}_{j=0}^k$.
\end{lemma}
    
Taking the union bound over $\{\phi_j\}_{j=0}^k$, we immediately obtain a uniform concentration bound.

\bprop[Uniform Concentration for the Estimation Error]\label{prop:dev_rev}
Let ${\cal P}$ be a set of $k$ prices and $\Pi(\cal P)$ be the family of all non-adaptive policies for price set $\cal P$.
Then, for any $\tau>0$, we have \begin{equation} \label{eqn:sup_error_tail_bound} 
\ho{P}\lb[ \sup_{{\bf t}\in \Pi({\cal P}) } \lb|{\rm Rev}({\bf t},\ \widehat \cI) - {\rm Rev}({\bf t},\ \cI)\rb| > \tau \rb] \le  2(k+1) \exp\lb( -\frac{\tau^2}{4 p_1^2n \sum_{j=1}^k  \frac 1{q(s_j)})}\rb).
\end{equation}
\eprop

\subsubsection{Concluding the Proof}
We combine all the above components to prove \cref{thm:regret_bound}, the main regret bound.
Intuitively, $\eta_2 =  |\E [{\rm Rev} (\widehat {\bf t},\ \cI )] - \E [{\rm Rev}({\bf t}^\star,\ \cI)]|$ is not too large because (i) the functions ${\rm Rev}(\cdot, \cI)$ and ${\rm Rev}(\cdot, \hat \cI)$ are uniformly close due to \cref{prop:dev_rev}, and (ii) $\widehat {\bf t}$ and ${\bf t^\star}$ maximize these two functions \resp. 
We formalize this idea as follows.

\begin{proposition}[Bounding $\eta_2$, the $k$-price setting] \label{lem:k_price_eta2_bound}
For all $\tau > 0$, we have
\[\eta_2 \le 4 \left( \tau + p_1n\cdot \ho{P}\lb[ \sup_{{\bf t}\in \Pi(\cal P)} \lb|{\rm Rev}({\bf t}, \widehat \cI) - {\rm Rev}({\bf t}, \cI)\rb| > \tau \rb] \right).\]
\end{proposition}
}

{ In \parti, if $s_j = s$ for all $j \in [k]$ and $\tau= p_1\sqrt{8 kn \log(nk) /q(s)}$, then the right-hand side of \cref{eqn:sup_error_tail_bound} becomes $O(n^{-2} k^{-1})$.
Therefore, by \cref{lem:k_price_eta2_bound},
\[\eta_2 = O(\tau + n \cdot n^{-2}k^{-1}) = O\lb(p_1 \sqrt {\frac{kn}{q(s)}}\rb).\]
Also, recall that $q(s)=O(\lam s)$ and from
\cref{lem:k_price_eta1_bound} that $\eta_1 = O(\lambda p_1 k n s)$, so we conclude that
\[{\rm Regret}\lb(X^{\rm LTE}, \cI\rb) = \widetilde{O}\lb(\lambda p_1 k n s + p_1\sqrt{ \frac{kn}{\lambda s}} \rb).\]
In \parti, for $s = \tilde\Theta(\lambda^{-1} (kn)^{-1/3})$, the above becomes $\tilde{O}\lb(p_1 (kn)^{2/3}\rb)$.
\hfill\Halmos
}


%% file: OR/Numerics.tex
We now numerically demonstrate the efficacy of our non-\adap\ and \adap\ policies. 
Specifically, we investigate (i) the empirical performance of our policies against policies designed for the stream model, and (ii) the impact of model misspecification, i.e., the loss due to mistakenly making decisions based on the stream model when the actual dynamics follow a pool model.
 


\subsection{Non-adaptive Setting} 

\subsubsection{The Robust Stream Policy.} \label{sec:robust_stream}
In \cref{sec:non_adap} we constructed a $1/k$-competitive non-\adap\ policy ${\bf t}^*_{\cal P}$ (see Theorem \ref{thm:non-adap}) for any price set ${\cal P}$. 
We will refer to this policy as the {\em Robust Pool} (RP) policy.
The same approach can also be used to find a robust policy for the stream model.

\Sps\ the seller believes that the true model is a stream model and has unlimited \inv.
In this setting, we can represent a policy as a \distr\ over $\cal P$. 
Formally, we denote by $t_i\in [0,1]$ the probability of choosing $p_i \in \mathcal{P}$ in each round.
The expected revenue per round is $ \sum_{i=1}^k v_i t_i (\sum_{j: 1\le j \le i} n_j)$, and the optimal detail-dependent policy has expected revenue $\max_{i \in [k]} \{v_i \sum_{j: 1\le j \le i} n_j\}$. 
A {\em robust stream} (RS) policy aims to find the $t_i$'s that maximizes the competitive ratio, i.e.,  
\[\min_{n_1, \dots, n_k} \frac{\sum_i v_i t_i \lb( \sum_{j\ge i} n_j\rb)}{\max_{i} \{v_i \cdot \sum_{j\ge i} n_j\}},\]
subject to $\sum_{i \in [k]} t_i = 1$ and $t_i\ge 0$. 
We derive a closed-form formula for the optimal stream policy, which we call the robust stream (RS) policy, in  \cref{apdx:robust_nonadap}.
In the following simulation, we will use this RP policy as a benchmark.

\subsubsection{Experimental Setup.}
We compare the average revenue of four policies:
(i) robust pool (RP) policy given \cref{sec:non_adap}, (ii) the robust stream (RS) policy we just derived above, and (iii) the optimal non-\adap\ (ON) policy. 

We consider two types of price sets.
For each $k = 2, \dots, 11$, we define the {\em uniform price sets}  $\{1 - (i - 1)/k\}_{i = 1}^k$ and {\em geometric price sets} $\{2^{-i}\}_{i = 0}^{k-1}$. 
For a fixed price set $\cal P$, we sample the valuations of $n = 100$ customers \unif ly from $\cal P$.
To facilitate comparisons, we also plot the average upper bound (UB) of the optimum given in Definition \ref{prop:ub}, averaged over these random instances. 
To highlight the impact of interaction rates, we consider $\lam = 1, 3, 5$ in each of the three subfigures. 

\subsubsection{Average Revenue}


\begin{figure}
     \centering
     \begin{subfigure}[b]{0.325\textwidth}
         \centering
         \includegraphics[width=\textwidth]{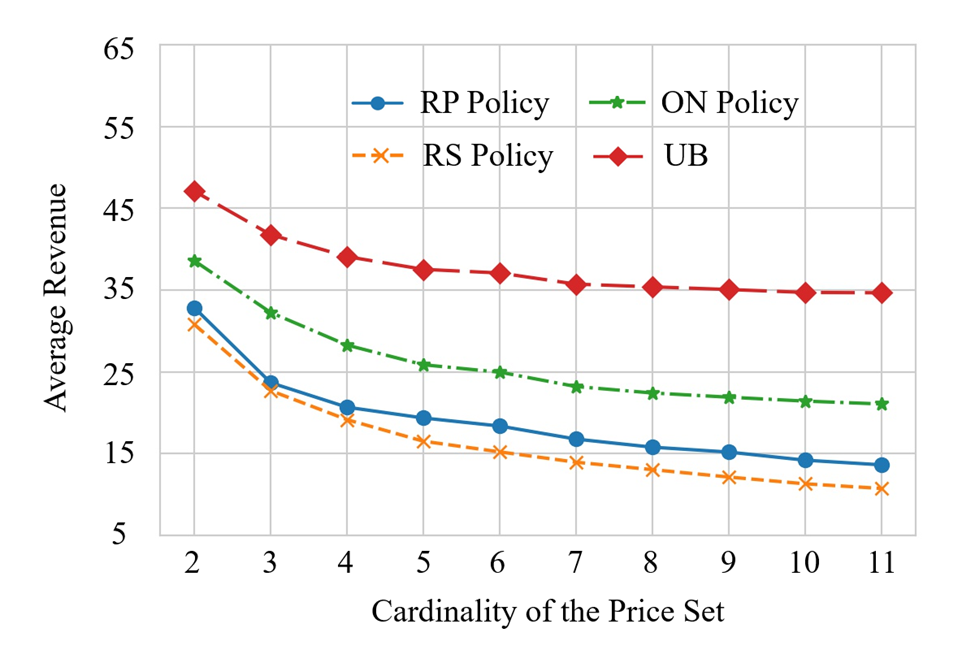}
     \end{subfigure}
     \hfill
     \begin{subfigure}[b]{0.325\textwidth}
         \centering
         \includegraphics[width=\textwidth]{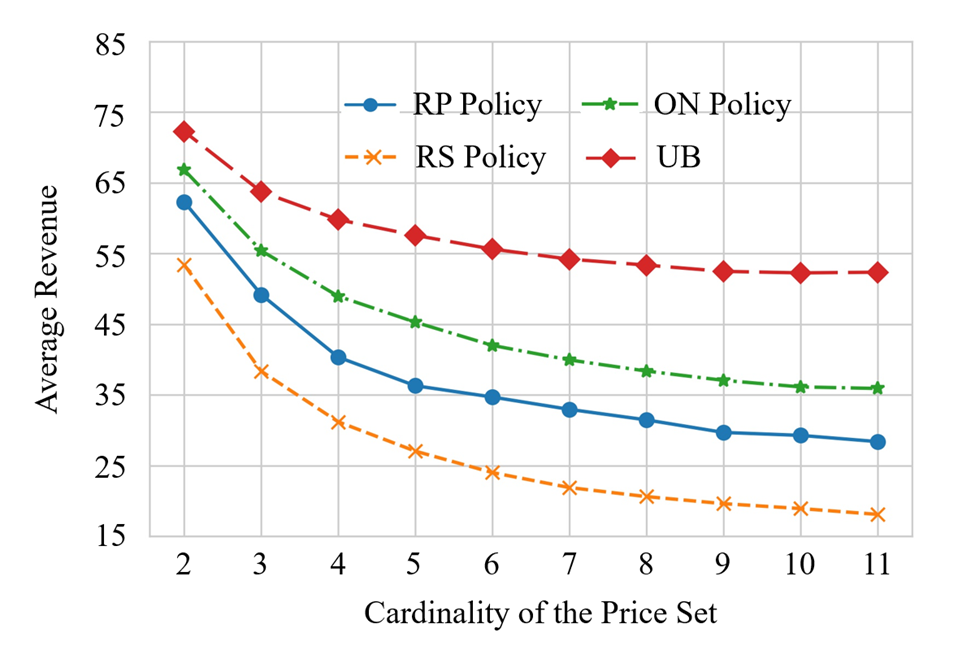}
     \end{subfigure}
     \hfill
     \begin{subfigure}[b]{0.325\textwidth}
         \centering
         \includegraphics[width=\textwidth]{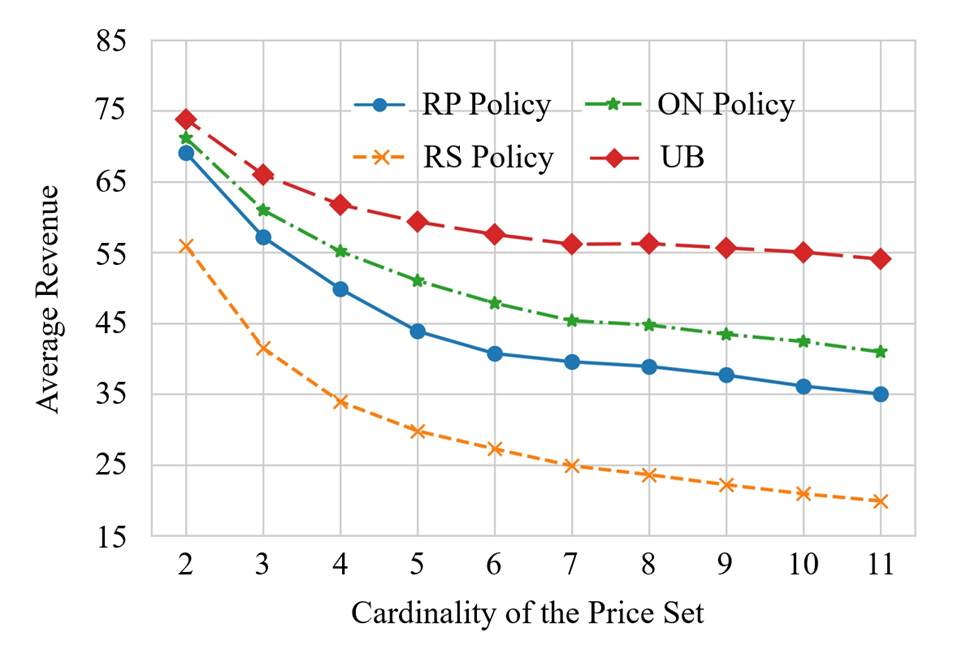}
     \end{subfigure}
     \caption{Average Revenue for Uniform Price Set. We plot the average revenue for uniform price sets with $k = 2, \dots, 11$. 
The interaction rate is chosen to be $1$, 3 and 5 in each of these three subfigures.}
 \label{fig:avg_rev_unif}
\end{figure}


\begin{figure}[ht]
     \centering
     \begin{subfigure}[b]{0.325\textwidth}
         \centering
         \includegraphics[width=\textwidth]{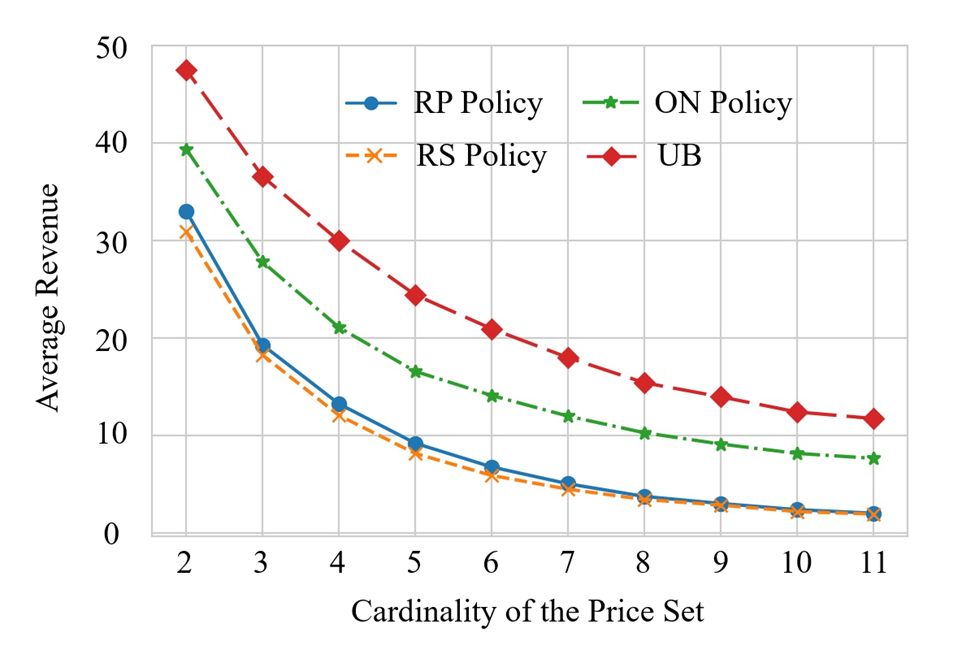}
     \end{subfigure}
     \hfill
     \begin{subfigure}[b]{0.325\textwidth}
         \centering
         \includegraphics[width=\textwidth]{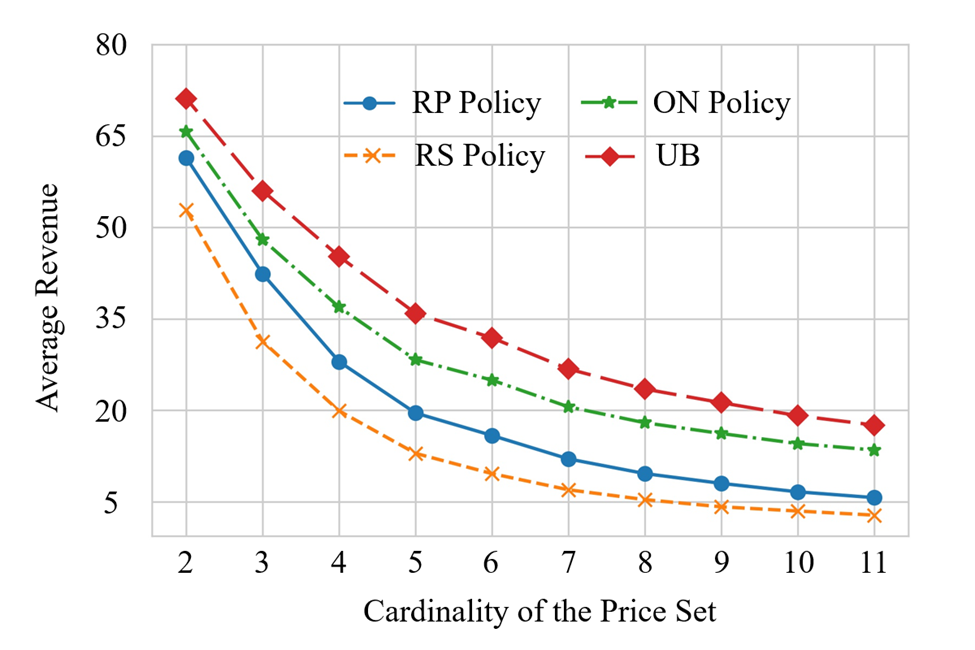}
     \end{subfigure}
     \hfill
     \begin{subfigure}[b]{0.325\textwidth}
         \centering
         \includegraphics[width=\textwidth]{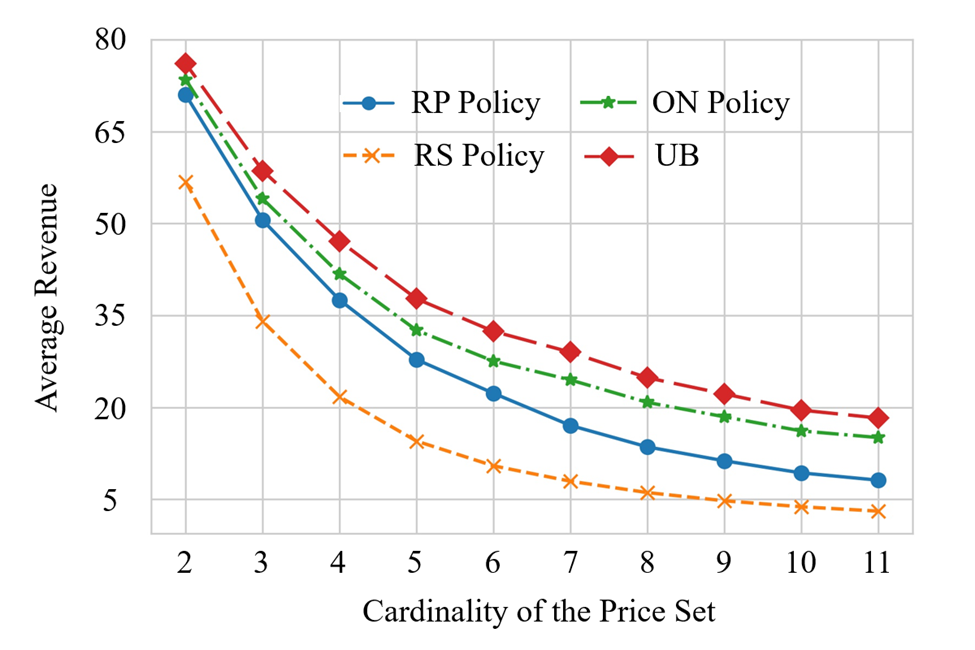}
     \end{subfigure}
\caption{Average Revenue for Geometric Price Set. We plot the average revenue for the geometric price sets for $k = 2, \dots, 11$. The interaction rate is chosen to be $1,3$ and $5$ in each of these three sub-figures.
}    
\label{fig:avg_rev_geo}
\end{figure}


 
We visualize the revenue averaged over $500$ runs for these two types of price sets in \cref{fig:avg_rev_unif,fig:avg_rev_geo}.
In each of these three subfigures, the RS policy substantially outperforms the {\em robust stream-based policy} consistently. 
Specifically, it is, on average, 18.0\%, 40.6\%, and 55.0\% higher when $\lambda = 1, 3, 5$ for the uniform price set. Similarly, for the geometric price set, the average revenue is 9.7\%, 65.2\%, and 103.7\% higher when $\lambda = 1, 3, 5$, respectively.
Through markdowns, the RP policy effectively extracts revenue from high-valuation customers before reducing the price. 
In contrast, the RS policy maintains a static price distribution over time. This potentially led high-valuation customers to pay lower prices and resulted in a revenue loss. 


Furthermore, the revenue difference between the RP and RS policies widens when $\lambda$ increases, suggesting that the loss due to model misspecification becomes more significant.
This is because when $\lam$ increases, the revenue of the RP policy increases faster than the RS policy.
In fact, a high $\lam$ helps the RP policy generate more revenue from high-valuation segments.
In contrast, its benefit on the RS policy is limited, since its revenue is proportional to $1-e^{-\lam}$, which is concave in $\lam$.

 
\subsubsection{Competitive Ratio}
In \cref{thm:non-adap}, we showed that the best possible competitive ratio achievable by non-\adap\ policies is $1/k$.
This analysis is worst-case in nature and may therefore be overly pessimistic.
Thus we investigate the empirical performance of the RP policy against the optimal policy.

\begin{figure}[ht]
     \centering
     \begin{subfigure}[b]{0.46\textwidth}
         \centering    \includegraphics[width=\textwidth]{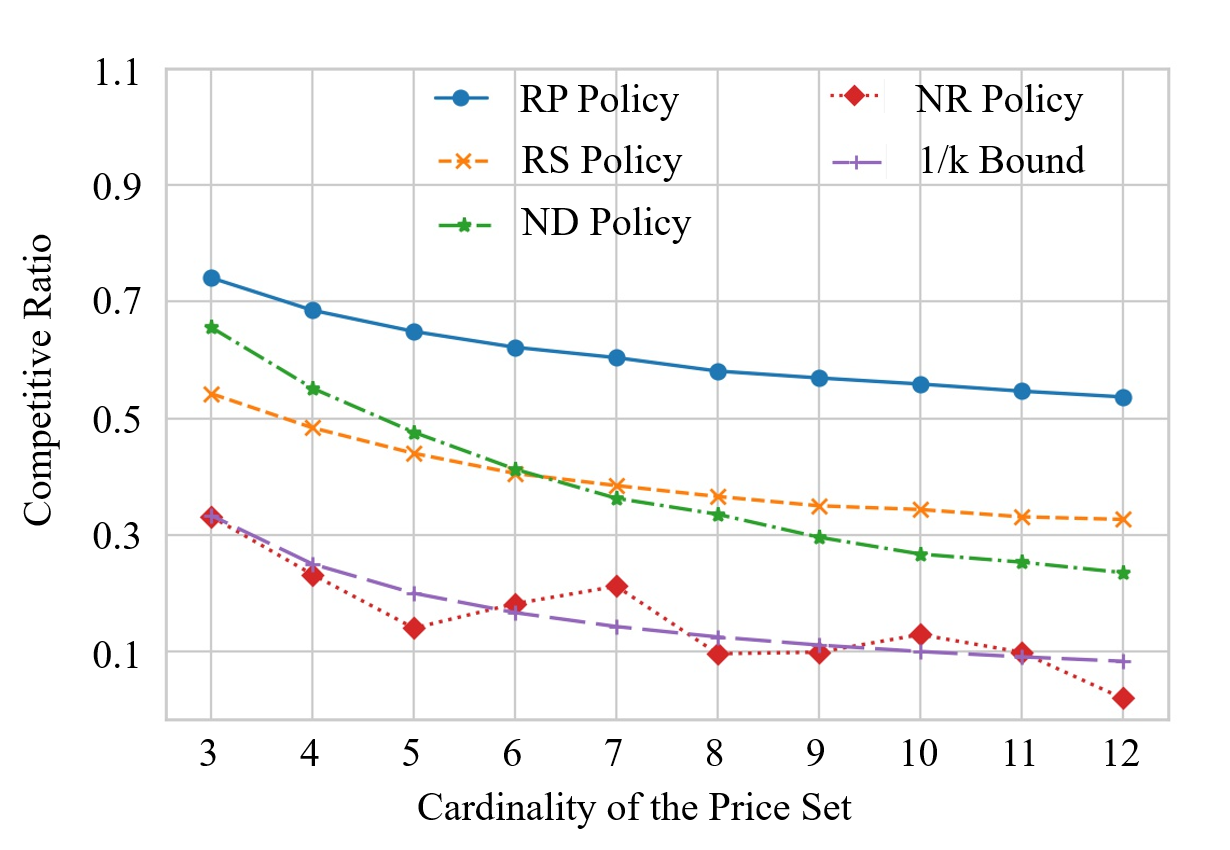}
     \end{subfigure}
     \hfill
     \begin{subfigure}[b]{0.46\textwidth}
         \centering      \includegraphics[width=\textwidth]{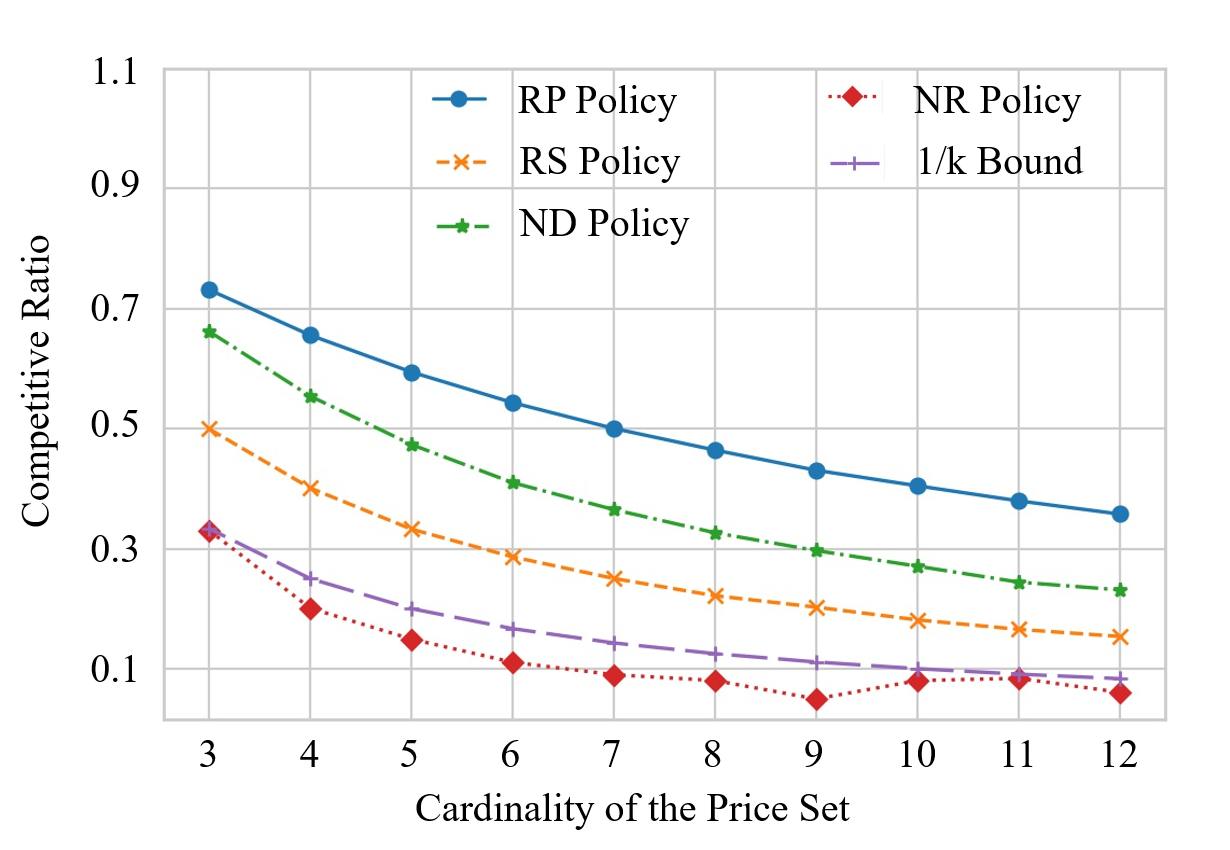}
     \end{subfigure}
\caption{Competitive ratio of RP and RS policy: For each $k = 3, \dots, 12$, we plot the (non-\adap) competitive ratios of the RP and RS policies as functions under the uniform and geometric price sets.
}
\label{fig:non_adap_ratio}
\end{figure}

To highlight the robustness of the RP policy, we compare it with the RS policy, as well as the NR and ND policies discussed in \cref{sec:impractical_policies}, as benchmarks.
For each price set, we run $100$ simulations for $n = 100$ customers.
The results are visualized in \cref{fig:non_adap_ratio} for the uniform and geometric price sets, \resp.

As evident from \cref{fig:non_adap_ratio}, the RP policy significantly outperforms all the other policies in terms of competitive ratio. 
In addition, consistent with our findings in the average revenue section, model misspecification incurs a substantial revenue loss in the worst-case scenario, especially for geometric price sets. 
This is because our policy initially provides a higher price, enabling us to generate more revenue, both on average and in the worst-case scenario.

Finally, we observe that the competitive ratio on the uniform price set is greater than the geometric price set.
This is consistent with our theoretical analysis in \cref{thm:non-adap}. 
In fact, substituting the expressions of $p_i$ into \cref{eqn:122723}, the competitive ratios for uniform and geometric price sets are $O(1/\log(\rho))$ and $O(1/k)$ where $\log \rho$, the log-ratio between the maximum and minimum prices, can be substantially lower than $k$.

\subsection{Adaptive Setting}
In this subsection, we investigate the impact of model misspecification within the adaptive setting.
{\new We compare our LTE policy with three classical bandit algorithms: {\em Explore-Then-Commit} (ETC) \citep{even2002pac}, {\em Upper Confidence Bound} (UCB) \citep{auer2002finite} and {\em Exponential-weight Algorithm for Exploration and Exploitation} (EXP3) \citep{auer1995gambling}.}

\begin{figure}[ht]
     \centering
     \begin{subfigure}[b]{0.43\textwidth}
         \centering
         \includegraphics[width=\textwidth]{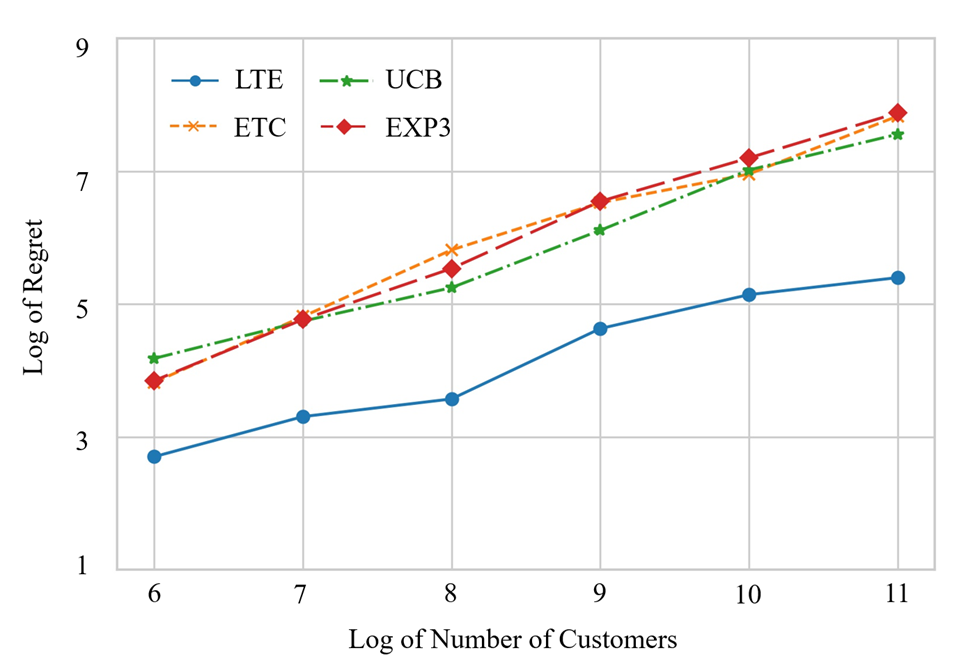}
     \end{subfigure}
     \hfill
     \begin{subfigure}[b]{0.43\textwidth}
         \centering
         \includegraphics[width=\textwidth]{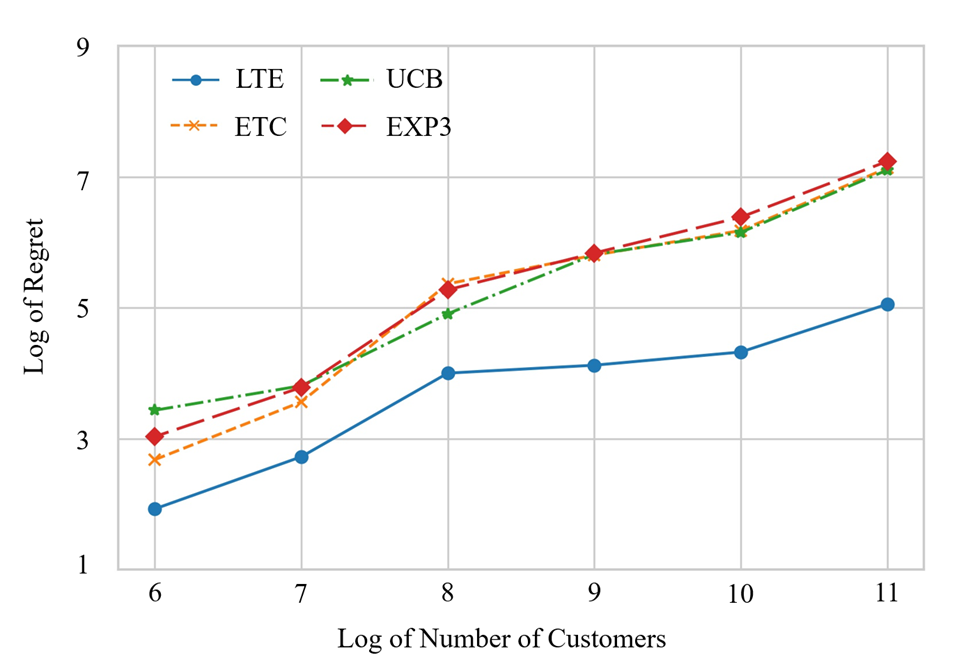}
     \end{subfigure}
\caption{Regret vs. Number of Customers: A Log-Log Plot. We plot the log regret v.s. log number of customers for uniform and geometric price sets. 
}
\label{fig:adap_regret}
\end{figure}


As in the previous analysis, we consider uniform and geometric price sets.
To understand how regret scales with the number $n$ of customers, we fix $k = 5$ and vary $n$.
Specifically, we set the valuation distribution as $\mathbf{q} = \left\{\frac{1}{2}, \frac{1}{4}, \frac{1}{8}, \frac{1}{16}, \frac{1}{16}\right\}$, with $n_i= n q_i$ for each $i$. 
To examine how regret increases with the growth of the total number of customers, we set $n = 2^{5+i}$ for $i = 1 \dots 6$. 
For each $n$, we calculated the average regret over $500$ simulations with interaction rate $\lambda=3$. 

As illustrated in \cref{fig:adap_regret}, our LTE policy significantly outperforms {\bf all} three policies designed for the stream model. 
The slope of our LTE policy on the uniform price set (left of \cref{fig:adap_regret}) is only $0.573$, whereas the slopes of the other three policies are $0.777$ (ETC), $0.702$ (UCB), and $0.813$ (EXP3).
Similarly, the slope of our LTE policy on the geometric price set (right of \cref{fig:adap_regret}) is $0.587$, while the slopes of the other three policies are $0.873$ (ETC), $0.751$ (UCB), and $0.840$ (EXP3), respectively.
These numerical results align with our $\Tilde{O}(n^{2/3})$ regret bound in \cref{thm:regret_bound} and show that classic MAB algorithms are subpar in our setting

%% file: OR/conclusion.tex
We introduce a novel single-item revenue maximization problem involving a {\bf pool} of unit-demand, non-strategic customers.
These customers interact repeatedly with a single seller, following independent Poisson processes, and purchase if the observed price is lower than their private valuation, after which they leave the market permanently.
We propose an algorithm that generates a non-adaptive, detail-free policy ensuring a $1/k$ fraction of the expected revenue compared to any policy. 
We also show that the policy is $(1 + \ln \rho)$-competitive, where $\rho$ is the ratio between the highest and lowest prices.
Furthermore, we present an adaptive LTE policy with s $\tilde O((kn)^{2/3})$ regret based on a novel debiasing technique. 
Additionally, we provide numerical simulations to demonstrate the effectiveness of our proposed non-adaptive and adaptive policies. 
We believe that this work paves the way for future research on pricing under the law of diminishing utility.
The following open questions may be of interest:
\begin{itemize}
    \item \textbf{General Discount Functions.} We established the first results dynamic pricing for the UDPM, i.e., $\psi\equiv 1$, which is the most basic special case of the pool model.  
    A natural question is, to what extent can we extend our results to more general discount functions $\psi(m)$?
    \item \textbf{Stronger Regret Guarantees.} In \cref{sec:adap} we gave a simple adaptive detail-free policy with a  $\tilde{O}((kn)^{2/3})$ regret against an optimal non-adaptive policy.
    Is there a policy with improved regret (in either $k$ or $n$)?
    \item \textbf{Regret Lower Bounds.} \cref{sec:adap} gives upper bounds on the regret for our LTE policy. 
    Can we establish minimax lower bounds on the regret for detail-free adaptive policies?  
    Standard lower bound approaches for MAB do not directly apply to our setting.
    \item \textbf{Understanding Adaptivity.} Finally, we remark that this paper has used non-adaptive policies as a benchmark class of policies.
    Naturally, one might ask if we can develop a better understanding of adaptive policies, even for the case of the UDPM?
\end{itemize}


%% file: OR/future.tex
%% file: OR/Appendix.tex
\newpage
\section{Omitted proofs from Section \ref{sec:known_demand}}

\subsection{Proof of \cref{prop:monotone}}

Suppose for contradiction the price \xulie\ is not non-increasing in some interval, say, the price is $X_s = p_L$ for $s \in [t-\eps,t]$ and increases to $ X_s = p_H$ for $X_s \in [t, t+\eps]$ where $\eps>0$. 
Let $d_L$ (resp. $d_H$) be the number of customers with a valuation greater or equal to than $p_L$ (resp. $p_H$) and are still in the market at time $t - \eps$.
Then, the expected revenue of $X$ in $[t - \eps, t + \eps)$ is
\begin{align*} A:= d_L p_L (1 - e^{-\lambda \eps}) + d_H p_H e^{-\lambda \eps}(1 - e^{-\lambda \eps}).
\end{align*}

Now, consider another policy $(X'_s)_{s\in [0,1]}$ where the prices in $[t,t+\eps]$ and $[t-\eps, t]$ are swapped. Formally,  we define $X'_s = X_s$ for $s \notin [t - \eps, t + \eps)$ and $X'_s = p_L$ for $s \in [t - \eps, t)$ and $X'_s = p_H$ for $s \in [t, t + \eps)$. 
Then, the expected revenue of $(X'_s)$ in $[t - \eps, t + \eps)$ is
\begin{align*}  B:= d_H p_H (1 - e^{-\lambda \eps})  + d_H p_L e^{-\lambda \eps}(1 - e^{-\lambda \eps}) + (d_L - d_H)p_L(1 - e^{-\lambda \eps}).
\end{align*}
Our statement follows by observing that \[ B- A = d_H p_H (1 - e^{-\lambda \eps})^2 - d_H p_L (1 - e^{-\lambda \eps})^2 = d_H (p_H - p_L) (1 - e^{-\lambda \eps})^2  \ge 0.\eqno\Halmos\]

\subsection{Proof of \cref{prop:concave}}\label{apdx:optimal_non_adap_grad}
{ We show that the Hessian of ${\rm Rev}({\bf t}, \mathcal{I})$ is negative semidefinite. 
To this end, we rewrite the    revenue function as 
\begin{align*}
    {\rm Rev}({\bf t}, \mathcal{I}) = \sum_{\ell = 1}^{k} n_\ell r_{\ell}(\textbf{t}) = \sum_{\ell=1}^k n_\ell, 
\end{align*}
where \[r_{\ell}(\textbf{t}) = p_j \left(e^{-\lambda \sum_{i = \ell}^{j - 1} t_i} - e^{-\lambda \sum_{i = \ell}^{j} t_i}\right),\]
is the expected revenue from type-$\ell$ customers.
Let $\mathcal{M}_{\ell}^{s}$ be a $k \times k$ matrix whose $(i,j)$-th entry is given by 
 $m_{i,j} = {\bf 1}(\ell \le i, j \le s)$. 
By \strfwd\ calculation, we can write the Hessian as a linear combination of the \mtxs\ $M_\ell^s$.

\begin{lemma}[Gradient and Hessian Matrix] \label{lem:r_grad_hessian}
For any $\ell \le j \le k$, we have 
\begin{align*}
    \frac{\partial r_{\ell}(\textbf{t})}{\partial t_j} = \sum_{s = j}^k \lambda \left(p_s - p_{s+1}\right)  e^{- \lambda \sum_{u = \ell}^{s} t_u},
\end{align*}
and 
\begin{align*}
    \frac{\partial^2 r_{\ell}(\textbf{t})}{\partial t_{i} \partial t_{j}}
    = \sum_{s = j}^k -\lambda^2 \left(p_s - p_{s+1}\right) e^{- \lambda \sum_{u = \ell}^{s} t_u}.
\end{align*}
Consequently, the Hessian matrix of $r_{\ell}(\cdot)$ is 
\begin{align*}
     \mathcal{H}_{\ell} = \sum_{j = \ell}^k -\lambda^2 \left( p_{j} - p_{j+1} \right) e^{-\lambda \sum_{i = \ell}^{j - 1} t_i} \mathcal{M}_{\ell}^j.
\end{align*}
\end{lemma}

\proof{Proof.}
Note that 
\begin{align*}
    \frac{\partial \left(e^{-\lambda \sum_{i = \ell}^{j}}\right)}{\partial t_s} = \begin{cases}
        - \lambda e^{-\lambda \sum_{i = \ell}^{j}}, & \text{ if } \ell \le s \le j,\\
        0, & \text{ Otherwise. }
    \end{cases}
\end{align*}
Therefore, for all $\ell \le j \le k $, the first order partial derivative of $r_{\ell}(\textbf{t})$ is
\begin{align*}
    \frac{\partial r_{\ell}(\textbf{t})}{\partial t_j} & =  p_j \lambda e^{- \lambda \sum_{u=\ell}^{j} t_u} + \sum_{s = j + 1}^{k} p_s\left( - \lambda e^{- \lambda \sum_{u=\ell}^{s-1} t_u} +  \lambda e^{- \lambda \sum_{u=\ell}^{s} t_u}  \right)\\
    & = \sum_{s = j}^k \lambda \left(p_s - p_{s+1}\right) \left(e^{- \lambda \sum_{u = \ell}^{s} t_u}\right),
\end{align*}
where the first equality follows from taking the derivative to each term of $r_{\ell}(\textbf{t})$, and the second equality follows from reorganizing the sum. For any other $j$, the first order partial derivative of $r_{\ell}(\textbf{t})$ is $ \frac{\partial r_{\ell}(\textbf{t})}{\partial t_j} = 0.$

Further, from the first-order partial derivative of $r_{\ell}(\textbf{t})$, we see that for any $\ell \le i \le j \le k$, the second order partial derivative of $r_{\ell}(\textbf{t})$ is
\begin{align*}
     \frac{\partial^2 r_{\ell}(\textbf{t})}{\partial t_{i} \partial t_{j}} = \frac{\partial^2 r_{\ell}(\textbf{t})}{\partial t_{j} \partial t_{i}} = \sum_{s = j}^k -\lambda^2 \left(p_s - p_{s+1}\right) \left(e^{- \lambda \sum_{u = \ell}^{s} t_u}\right).
\end{align*}
For any other $i, j$, the second order partial derivative of $r_{\ell}(\textbf{t})$ is $ \frac{\partial^2 r_{\ell}(\textbf{t})}{\partial t_{i} \partial t_{j}} = 0$. With the definition of $\mathcal{M}_{\ell}^s$, the Hessian matrix of $r_{\ell}(\textbf{t})$ can be reformulated as
\begin{align*}
     \mathcal{H}_{\ell} = \sum_{j = \ell}^k -\left( p_{j} - p_{j+1} \right) e^{-\sum_{i = \ell}^{j - 1} t_i} \mathcal{M}_{\ell}^j. 
\end{align*}
Since $\mathcal{M}_{\ell}^j$ is positive semidefinite (all-ones matrix is positive semi-definite) and $p_{j} \ge p_{j+1}$ for any $j$, the Hessian matrix $\mathcal{H}_{\ell}$ is negative semidefinite, i.e., the function $r_{\ell}(\textbf{t})$ is concave for any $\ell$.
\hfill\Halmos
}

\section{Omitted proofs from Section \ref{sec:non_adap}}

\subsection{Proof of \cref{thm:non_adap_upper}: Upper Bound on the Competitive Ratio}\label{apdx:UB}
{ We will prove the result for deterministic policies and in the end explain why it also holds for randomized policy.
We construct the instance $\cI_\eps$ as follows. The prices are given by a geometric sequence $p_{i+1} = p_i/a$ where $a>0$ is a constant to be specified later.
The valuation \distr\ $(n_i)_{i\in [k]}$ is given by $n_i = n \cdot {\bf 1}(i=i^*)$ where \[i^* = \arg\min_{i \in [k]} \sum_{j \ge i} \frac{p_j}{p_i} t_j.\]

We aim to that any non-\adap\ policy $\pi$ has a low competitive ratio on $\cI_\eps$, if $\lambda$ is sufficiently small.
Note that since all customers have the same valuation, we have ${\rm OPT}( \mathcal{I}_\eps) = p_{i^*} (1 - e^{-\lambda})$. It then follows that 
\begin{align}\label{eqn:030524}
\frac{{\rm Rev}({\bf t}, \cI_\eps)}{{\rm OPT}(\cI_\eps)} 
& = \frac{\sum_{j \in [k]: j \ge i^*}   p_j \left(e^{-\lambda \sum_{\ell = i}^{j-1} t_{\ell}} - e^{-\lambda \sum_{\ell = i}^{j} t_{\ell}}\right)}{p_{i^*} (1 - e^{-\lambda})} \notag\\
& = \sum_{j \in [k]: j \ge i^*}  \frac{p_j}{p_{i^*}} \left(\sum_{\ell = i}^{j} t_{\ell} - \sum_{\ell = i}^{j-1} t_{\ell} + \frac{\lambda}{2} \left(\sum_{\ell = i}^{j} t_{\ell} - \sum_{\ell = i}^{j-1} t_{\ell}\right) \left(1 - \left( \sum_{\ell = i}^{j} t_{\ell} + \sum_{\ell = i}^{j-1} t_{\ell}\right)\right) + O(\lambda^2)\right) \notag\\
& = \frac{ \sum_{j \in [k]: j \ge i^*}   p_j t_j }{p_{i^*}} + \frac{\lambda}{2} \sum_{j \ge i^*} \frac{p_j}{p_{i^*}} t_j \left(1 - \left(\sum_{\ell = i}^{j} t_{\ell} + \sum_{\ell = i}^{j-1} t_{\ell} \right)\right) + O(\lambda^2), 
\end{align}
where the second equality follows from Taylor's expansion in $\lambda$ since for any constant $c_1$ and $c_2$
\begin{align*}
    \frac{e^{- c_1 \lambda} - e^{-c_2 \lambda}}{1 - e^{-\lambda}} = \left(c_2 - c_1\right) + \frac{\lambda}{2}\left(c_2 - c_1\right)\left(1 - (c_1 + c_2)\right) + O(\lambda^2),
\end{align*}
and the last equality follows from simplification.

To proceed with the analysis, we next bound \cref{eqn:030524} in terms of ${\rm Rev}'(\cdot)$ and ${\rm OPT}'(\cdot )$, the linearized revenue of policy $\textbf{t}$ and optimal revenue, where ${\rm Rev}'(\textbf{t}, \mathcal{I}_\eps) = \sum_{j \in [k]: j \ge i^*}  \lambda p_j t_j$ and ${\rm OPT}'(\mathcal{I}_\eps) = \lambda p_{i^*}$.
Note by the definition of $i^*$ that $p_j / p_{i^*} \le 1$ for all $j\in [k]$, and
\[t_j \left(1 - \left(\sum_{\ell = i}^{j} t_{\ell} + \sum_{\ell = i}^{j-1} t_{\ell} \right)\right) \le t_j \le 1 \quad \text{for all } j \ge i^*.\] 
Therefore, for $\lam \le \eps/k$,  we have 
\begin{align}\label{eqn:030624}
\eqref{eqn:030524} \le \frac{{\rm Rev'}({\bf t}, \mathcal{I}_\eps)}{{\rm OPT'}(\mathcal{I}_\eps)} + \frac{\lambda k}2 + O(\lambda^2)
\le \frac{{\rm Rev'}({\bf t}, \mathcal{I}_\eps)}{{\rm OPT'}(\mathcal{I}_\eps)} + \frac\eps 2.
\end{align} 

Next, we consider the maximization of $\frac{{\rm Rev'}({\bf t}, \mathcal{I}_\eps)}{{\rm OPT'}(\mathcal{I}_\eps)}$. By the definition of $i^*$, the ratio is maximized at 
\begin{align}\label{eq:ratio_eq}
    \frac{\sum_{j \in [k]: j \ge i} p_j  t_j }{p_i} = \frac{\sum_{j \in [k]: j \ge i^* } p_j  t_j }{p_{i^*}}, \quad \forall i, i^* \in [k].
\end{align}
The solution of \cref{eq:ratio_eq} $(\textbf{t}^\star)$ satisfies
\begin{align} \label{eqn:010423}
    t_k^\star = \frac{1}{k - \sum_{1\le i \le k-1} \frac{p_{i+1}}{p_{i}}}, \quad t_{j}^\star = \left(1 - \frac{p_{i+1}}{p_{i}} \right) t_k^\star, \quad \forall j < k.
\end{align}
Picking $i^* = k$ will yield the competitive ratio as $ \frac{{\rm Rev'}({\bf t}, \mathcal{I}_\eps)}{{\rm OPT'}(\mathcal{I}_\eps)} = t_k^* = \frac{1}{k - \frac{k-1}{a}}$.

To conclude, we choose \[a= \frac{k - 1}{k - \frac 1{\frac 1k + \frac \eps 2}},\]
This is a valid choice since for any $\eps>0$ and integer $k\ge 1$ we have $a>1$. Then,
\begin{align}\label{eqn:030624b}
\frac{{\rm Rev'}({\bf t}, \mathcal{I}_\eps)}{{\rm OPT'}(\mathcal{I}_\eps)} = \frac{1}{k - \frac{k - 1}{a}} = \frac 1k + \frac{\eps}{2}.
\end{align}
Combining \cref{{eqn:030624}} and \cref{eqn:030624b}, we conclude that
\[ \frac{{\rm Rev}({\bf t}, \cI_\eps)}{{\rm OPT}(\cI_\eps)} \le \frac 1k + \eps.\eqno \Halmos\]

{\bf Extending to randomized policies.} 
Due to the concavity of the revenue function (\cref{prop:concave}) and Jensen's inequality, any randomized non-adaptive detail-free policy can be converted to a deterministic one for which the competitive ratio can only increase. 
More formally, if $D$ is any distribution over sequences ${\bf t} = (t_j)_{j \in [k]}$ with $\sum_{j=1}^k t_j = 1$, then we have
$\E_{{\bf t} \sim D}[{\rm Rev}({\bf t}, {\cal I})] \leq {\rm Rev}( \E_{{\bf t} \sim D}[{\bf t}], {\cal I})$ 
for any instance ${\cal I}$
due to \cref{prop:concave} and Jensen's inequality.
}

\subsection{Proof of \cref{prop:naive_randomized}}
Recall that NR commits to a price uniformly at random.  
In other words, we sample $j$ uniformly at random from $[k]$, and then set the policy to be ${\bf t}$, where we set $t_j = 1$ and $t_{i} = 0$ for all $i \neq j$.
From \cref{prop:exp_rev}, we can see that the expected revenue of this policy is
\[
\E[{\rm Rev}({\bf t}, \mathcal{I})] = \frac{1}{k} \sum_{j=1}^k p_j\sum_{\ell: j \geq \ell} n_\ell (1- e^{-\lambda}) \geq \frac{1}{k} (1-e^{-\lambda}) \sum_{j=1}^k p_j n_j = \frac{1}{k} {\rm UB}(\cal I),
\]
where ${\rm UB}(\cal I)$ is an upper bound of the optimal achievable revenue. Therefore, the NR policy is $1/k$-competitive. 
\hfill\Halmos

\subsection{Proof of \cref{prop:naive_deterministic}}
This follows directly from \cref{prop:naive_randomized}, \cref{prop:concave}, and Jensen's inequality. 
To see this, let ${\bf t}$ be a random policy sampled according to the NR policy, then we have $\E[{\bf t}] = (1/k, 1/k,\ldots, 1/k)$, which is exactly the ND policy. Thus, the expected revenue of the ND policy satisfies
\begin{align*}
    \E\left[{\rm Rev}(\E[{\bf t}], \mathcal{I})\right] \ge \E[{\rm Rev}({\bf t}, \mathcal{I})] \ge \frac{1}{k} {\rm UB}(\cal I),
\end{align*}
which completes the proof.
\hfill\Halmos

\subsection{Proof of \cref{prop:ub}}
{ Recall that in the UDPM, it is w.l.o.g. to assume that the valuation of each customer is from the price set $\cal P$.
\Sps\ customer $i$ has valuation $p_j\in \cal P$. 
Then, for any non-\adap\ policy $X$, the revenue $R_i$ from this customer \sats\
\begin{align*}
\ho{E} [R_i] = \int_{0}^{1} X_s {\bf 1} (X_s \le p_j)\ e^{-\lambda s} ds \le \int_{0}^{1} p_j e^{-\lambda s} ds = p_j(1 - e^{-\lambda}).
\end{align*}
Therefore,
\[ {\rm Rev}(X, \cI) = \sum_{i\in [n]} \ho{E} [R_i] \le \sum_{j\in [k]} n_j \cdot p_j (1 - e^{-\lambda}).
\eqno\Halmos\]
}

\subsection{Proof of \cref{thm:non-adap}: the $1/k$-Competitive Ratio} \label{apdx:non_adap}
{
\subsubsection{Linearization}
Recall from Proposition \ref{prop:ub} that ${\rm UB}(\mathcal{I})$ is an upper bound on the expected revenue of any detail-dependent adaptive policy for any instance $\cal I$.
Thus, if ${\rm OPT}(\mathcal{I})$ is replaced by ${\rm UB}(\mathcal{I})$, the objective function of (P0) becomes lower, and hence any $\alpha$-\apxn\ for (P1) is also an $\alpha$-\apxn\ for (P0).
Formally, consider
\begin{align*}
&\text{(P1)} 
\quad \max_{t_1,\dots,t_k} \min_{n_1,\dots,n_k}
\frac{{\rm Rev}((t_i)_{i\in [k]}, \mathcal{I})}{{\rm UB}(\mathcal{I})},\\
& \text{such that } \quad  \sum_{i=1}^k t_j = 1, \quad t_j \ge 0, \quad \forall j \in [k]. 
\end{align*}

Unfortunately, (P1) is still not readily solvable since ${\rm Rev}({\bf t}, \mathcal{I})$ and ${\rm UB}(\mathcal{I})$ are both non-linear (in $t_i$'s).
To address this, we consider {\em linear surrogates} motivated by Taylor's expansion.

\bdefn[Linear Surrogate]
For any \ins\ $\mathcal{I} =(\lambda, \{n_i\}_{i\in [k]}, \{p_i\}_{i\in [k]})$ and  non-\adap\ policy ${\bf t} = (t_i)$, we define the linear surrogate of ${\rm UB(\mathcal{I})}$ and ${\rm Rev(\pi, \mathcal{I})}$ as
\begin{align*}
{\rm UB}'(\mathcal{I}) &:=\sum_{i \in [k]} n_i p_i \lambda,\\ 
{\rm Rev}'(\pi, \mathcal{I}) &:= \sum_{i \in [k]} n_i\sum_{j \in [k]: j \ge i} \lambda p_j  t_j.
\end{align*}
\edefn

We show that this linearization only decreases the \obj\ in (P1). 
Thus, a lower bound on the linearized max-min program implies a lower bound on (P1).

\begin{lemma}[Linearization Reduces the Objective]
\label{lem:linear_approx}
For any instance $\mathcal{I} = (\lambda, \{n_i\}_{i=1}^k,\{p_i\}_{i=1}^k)$ and non-\adap\ policy ${\bf t}$, we have 
\[\frac{{\rm Rev}({\bf t}, \mathcal{I})}{{\rm UB}(\mathcal{I})} \ge \frac{{\rm Rev'}({\bf t}, \mathcal{I})}{{\rm UB'}(\mathcal{I})}.\] 
\end{lemma}
\proof{Proof.} Recall that for any instance $\mathcal{I} = (\lambda,\{n_i\}_{i=1}^k,\{p_i\}_{i=1}^k)$ and non-\adap\ policy $\bf t$, the expected revenue is 
\begin{align*}
    {\rm Rev}({\bf t}, \mathcal{I}) = \sum_{\ell=1}^k n_\ell \sum_{j:\ell\le j \le k} p_j  \left(e^{-\lambda \sum_{i = \ell}^{j - 1} t_i} - e^{-\lambda \sum_{i = \ell}^{j} t_i}\right).
\end{align*}
Recall
from \cref{lem:r_grad_hessian} that each function $r_{\ell}(\textbf{t}) = \sum_{j:\ell\le j \le k} p_j  \left(e^{-\lambda \sum_{i = \ell}^{j - 1} t_i} - e^{-\lambda \sum_{i = \ell}^{j} t_i}\right)$ is concave.
Let $\textbf{1}_j$ be the vector with $j$-th element to be 1 and all the other elements to be 0, then the value of $ r_{\ell}(\textbf{1}_j)$ is 
\begin{align*}
    r_{\ell}(\textbf{1}_j) = \begin{cases}
        p_j (1 - e^{\lambda}), & \text{ for all } \ell \le j \le k,\\
        0, & \text{ otherwise. }
    \end{cases}
\end{align*}
Since $\sum_{i=1}^k t_i = 1$, by Jensen's inequality, we have
\begin{align*}
    r_{\ell}(\textbf{t}) \ge \sum_{j = 1}^k t_j r_{\ell}\left({\textbf{1}_j}\right).
\end{align*}
Namely, we have the dominance 
\begin{align*}
    r_{\ell}(\textbf{t}) = \sum_{j:\ell\le j \le k} p_j  \left(e^{-\lambda \sum_{i = \ell}^{j - 1} t_i} - e^{-\lambda \sum_{i = \ell}^{j} t_i}\right) \ge \sum_{j:\ell\le j \le k} p_j  t_j \left(1 - e^{\lambda}\right).
\end{align*}
Then, summing over all $\ell \in [k]$, we obtain
\begin{align}\label{eq:rev_dominance}
    \sum_{\ell=1}^k n_\ell \sum_{j:\ell\le j \le k} p_j  \left(e^{-\lambda \sum_{i = \ell}^{j - 1} t_i} - e^{-\lambda \sum_{i = \ell}^{j} t_i}\right) \ge \sum_{\ell=1}^k n_\ell \sum_{j:\ell\le j \le k} p_j  t_j \left(1 - e^{\lambda}\right).
\end{align}
Dividing both sides of \cref{eq:rev_dominance} by $\sum_{i = 1}^k n_i p_i (1 - e^{-\lambda})$, we obtain
\begin{align*}
    \frac{\sum_{\ell=1}^k n_\ell \sum_{j:\ell\le j \le k} p_j  \left(e^{-\lambda \sum_{i = \ell}^{j - 1} t_i} - e^{-\lambda \sum_{i = \ell}^{j} t_i}\right)}{\sum_{i = 1}^k n_i p_i (1 - e^{-\lambda})} \ge \frac{\sum_{\ell=1}^k n_\ell \sum_{j:\ell\le j \le k} p_j  t_j}{\sum_{i = 1}^k n_i p_i},
\end{align*}
i.e., 
\begin{align*}
    \frac{{\rm Rev}({\bf t}, \mathcal{I})}{{\rm UB}(\mathcal{I})} \ge \frac{{\rm Rev'}({\bf t}, \mathcal{I})}{{\rm UB'}(\mathcal{I})}.\tag*\Halmos
\end{align*}

 


\subsubsection{Reducing to an LP}
With the result of \cref{lem:linear_approx}, now we further simplify the max-min program (P1) by replacing the \obj\ with the ratio between the linear surrogates, leading to 
\begin{align}\label{eq:P2}
\text{(P2)} \quad
& \max_{t_1,\dots,t_k} \min_{n_1,\dots,n_k} \frac{\sum_{i \in [k]} n_i\sum_{j \in [k]: j \ge i} p_j  t_j }{\sum_i n_i p_i},
\nonumber \\
& \text{such that } \quad \sum_{j=1}^{k} t_j = 1, \quad t_j \ge 0, \quad \forall j \in [k]. 
\end{align}
Next, we construct an optimal \sln\ to the (P2) by reducing it to a linear program (LP).
Observe that the inner minimum is always achieved by a binary vector with exactly one non-zero entry. More precisely, it is given by $n_i = n \cdot {\bf 1}(i=i^\star)$ where 
\[i^\star = \arg
\min \lb\{\frac{\sum_{j=i}^k p_j t_j}{p_i}:i\in [k]\rb\}.\]
(For simplicity, we assume $i^\star$ is unique; apparently, this is not essential to the analysis.)
Thus, (P2) can be reformulated as
\begin{align}\label{eq:P2_reformed}
    \max_{(t_i), c} \quad & c \nonumber \\
    \text{subject to }  \quad & c \le \frac{\sum_{j \in [k]: j \ge i} p_j  t_j }{p_i}, \, \forall i \in [k], \nonumber\\
    & \sum_{j=1}^n t_j = 1, \quad t_j \ge 0 \quad \forall j \in [k].
\end{align}
The optimum is achieved when all the inequalities are binding. In this case, the optimal solution $(\textbf{t}^\star)$ satisfies
\begin{align} \label{eqn:010423}
    t_{j}^\star = \left(1 - \frac{p_{i+1}}{p_{i}} \right) t_k^\star, \quad \forall j < k,
\end{align}
which solves to \[t_k^\star = \frac{1}{k - \sum_{1\le i \le k-1} \frac{p_{i+1}}{p_{i}}}.\]
Finding $t^\star_j$ for $j < k$ can be done with backward substitution using equation~\eqref{eqn:010423}. Since the ratio $\left(\sum_{j=i}^k p_j t_j^*\right)/p_i$ is the same for all $i \in [k]$, the resulting performance guarantee is given by \[{\rm CR}_{\rm NA}(p_1,\ldots,p_k) = \frac{p_k t_k^*}{p_k} = t_k^\star =  \frac{1}{k - \sum_{i=1}^{k-1} p_{i+1}/p_{i}}.\eqno\Halmos\]
}

\subsection{Proof of \cref{thm:conti_P}} \label{apdx:aspect_ratio}
{  
Since we have bounded the competitive ratio in the main body \cref{thm:conti_P}, here we focus on deriving the closed-form formula \cref{eqn:031424} for the optimal non-\adap\ policy.  
To this end, recall from \cref{def:non_adap} in the main body that given any {\em finite} price set $(p_i)_{i\in [k]}$, the optimal policy \sats\ \[t_k^* = \frac 1{k - \sum_{1\le i \le k-1} \frac{p_{i+1}}{p_{i}}}.\]
Since the competitive ratio improves (i.e., becomes larger) as $\eps$ decreases, we let $\eps$ tend to $0$, and obtain
\begin{align*}
    \lim_{\eps \rightarrow 0} t_k^* & = \lim_{\eps \rightarrow 0} \frac{1}{k - \sum_{1\le i \le k-1} \frac{p_{i+1}}{p_{i}}}\\
    & = \lim_{\eps \rightarrow 0}\frac{1}{k - \sum_{1\le i \le k-1} \frac{1}{1 + \eps}}\\
    & = \lim_{\eps \rightarrow 0} \frac{1 + \eps}{k(1 + \eps) - (k-1)}\\
    & = \lim_{\eps \rightarrow 0} \frac{1 + \eps}{1 + k \eps}\\
    & = \lim_{\eps \rightarrow 0} \frac{1 + \eps}{1 + \eps \log_{1 + \eps} \frac {p_{\max}}{p_{\min}} }\\
    & = \frac{1}{1 + \ln \frac {p_{\max}}{p_{\min}} },
\end{align*}
where the second equality follows since $p_{i+1}/p_i = 1/(1 + \eps)$. 
Recall that by definition, $t_k^*$ is the amount of time that the policy selects $p_{\min}$, so  we have $X_t = p_{\min}$ when $t$ \sats\ $1- \frac{1}{1 + \ln( p_{\max}/p_{\min})} \le t \le 1.$ 
  
Now, consider $t < 1 - \frac{1}{1 + \ln(p_{\max}/p_{\min})}$. 
Recall from \cref{eqn:031424} that
\begin{align*}
    t_i^* = \left(1 - \frac{p_{i+1}}{p_i}\right)t_k^* = \left(1 - \frac{1}{1 + \eps}\right) t_k^* = \frac{\eps}{1 + \eps} t_k^*.  
\end{align*}
Fix some integer $i$ and consider the time \[t = \sum_{j=1}^{i-1} t_{j}^* = \left(\frac{(i-1)\eps}{1 + \eps}\right) t_k^*\] when the policy starts to select $p_i$. 
Rearranging, we have $i = i(t) = 1 + \frac{t(1 + \eps)}{\eps t_k^*}$.
Since $p_{i} = p_{\max}(1+\eps)^{-(i-1)}$, we have 
\begin{align*}
    X_{t,\eps} =  p_{i(t)} = p_{\max}(1+\eps)^{- \frac{t(1 + \eps)}{\eps t_k^*}}.
\end{align*}
Therefore,  
\begin{align*}
X_t = \lim_{t\rar \infty} X_{t,\eps} = \lim_{\eps \rightarrow 0}  p_{\max}(1+\eps)^{- \frac{t(1 + \eps)}{\eps t_k^*}} = p_{\max} e^{-t \left(1 + \ln \frac {p_{\max}}{p_{\min}}\right)},
\end{align*}
where the equality follows since \[\lim_{\eps \rar 0} \ (1+\eps)^{- \frac{t(1 + \eps)}{\eps}} p_{\max} = e^{-t} \quad \text{and} \quad \lim_{\eps \rightarrow 0} t_k^* = \frac 1{1+\ln(p_{\max}/p_{\min})}\eqno\Halmos.\]
} 
\section{Robust Non-\adap\ Policy for the Stream Model}
\label{apdx:robust_nonadap}
\begin{theorem}[Competitive Ratio, Stream Model] For any $ k \ge 1$ and detail-free non-\adap\ policy $X$,
there is a stream instance $\mathcal{I} \in \mathcal{F}_k$ such that
\[\frac{{\rm Rev}(X, \mathcal{I})}{{\rm OPT}(\mathcal{I})}\le \frac 1k.\]
\end{theorem}
\proof{Proof.}
    For the stream model, we assume at each round, one customer arrives with the valuation randomly drawn from $\mathcal{P} = \{p_1>\dots > p_k\}$ with probability $\{q_1, \dots, q_k\}$ and $\sum_{i = 1}^k q_i = 1$. We define a non-\adap\ policy for the stream model is $X = \{t_1, \dots t_k\}$ and $\sum_{i = 1}^k t_i = 1$ where $t_i$ denotes the probability the seller chooses price $p_i \in \mathcal{P}$ in each round. Therefore, the competitive ratio of a non-\adap\ policy $X$ for the stream model is defined as 
    \[{\rm CR}(X, \mathcal{F}) := \inf_{\mathcal{I}\in \cal F} \frac{{\rm Rev}(X, \mathcal{I})}{{\rm OPT}(\mathcal{I})} = \inf_{\{q_1, \dots, q_k\} \in \cal F} \frac{\sum_{i = 1}^k t_i p_i \left(\sum_{j \le i} q_j\right)}{\max_{i} p_i \left(\sum_{j \le i} q_j\right)}.\]
    Note $\sum_{j \le i} q_j$ is the probability that customers arrive with a valuation larger than the price $p_i$.

    To show $\max_{X} {\rm CR}(X, \mathcal{F}) = \frac{1}{k}$, we first show that for any $X$, the competitive ratio ${\rm CR}(X, \mathcal{F})$ is minimized at $q_i = 1$ for some $i \in [k]$. If not, suppose the denominator $p_i \left(\sum_{j \le i} q_j\right)$ is maximized at $i^*$, and there exist $i < i^*$ such that $q_i > 0$.
    { We can construct a new instance $\{q'_i\}_{i=1}^{k}$ which will give us a smaller ratio. 
    Specifically, let $q'_{i^*} = q_i^* + q_i$, $q'_i = 0$ and $q'_j = q_j$ for all $j \in [k] \backslash \{i, i^*\}$. 
    Now we have $p_{i^*} \sum_{j = 1}^{i^*} q'_{j} = p_{i^*} \sum_{j = 1}^{i^*} q_{j}$ and $\sum_{i = 1}^k t_i p_i \left(\sum_{j \le i} q'_j\right) < \sum_{i = 1}^k t_i p_i \left(\sum_{j \le i} q_j\right) $ since customer's valuation is more likely to be lower than before. Then, the ratio for instance $\{q_i'\}_{i=1}^{k}$ will be smaller. Therefore, the optimal $\{q_1, \dots, q_k\}$ satisfies $q_i = 0$ for any $i < i^*$. 
    }

    Moreover, if there exists $i > i^*$ such that $q_i > 0$, let $q'_{i^*} = q_i^* + q_i$, $q'_i = 0$ and $q'_j = q_j$ for all $j \in [k] \backslash \{i, i^*\}$. Then, the competitive ratio is
    \begin{align*}
        \frac{\sum_{j = 1}^k t_j p_j \left(\sum_{j' \le j} q'_{j'}\right)}{p_{i^*} \left(q_{i^*} + q_{i}\right)} = \frac{\sum_{j = 1}^k t_j p_j \left(\sum_{j' \le j} q_{j'}\right) + \sum_{j = i^*}^{i - 1} t_j p_j q_i }{p_{i^*} \left(q_{i^*} + q_{i}\right)},
    \end{align*}
where $\sum_{j = i^*}^{i - 1} t_j p_j q_j$ is the extra revenue since we set $q'_{i^*} = q_{i^*} + q_i$. Now to show the competitive ratio will decrease, i.e, 
\begin{align*}
    \frac{\sum_{j = 1}^k t_j p_j \left(\sum_{j' \le j} q_{j'}\right) + \sum_{j = i^*}^{i - 1} t_j p_j q_i }{p_{i^*} \left(q_{i^*} + q_{i}\right)} \le \frac{\sum_{j = 1}^k t_j p_j \left(\sum_{j' \le j} q_{j'}\right)}{p_{i^*} q_{i^*}},
\end{align*}
we only need to show that
\begin{align*}
    \left(\sum_{j = 1}^k t_j p_j \left(\sum_{j' \le j} q_{j'}\right) + \sum_{j = i^*}^{i - 1} t_j p_j q_i \right) \left(p_{i^*} q_{i^*}\right) \le \left(\sum_{j = 1}^k t_j p_j \left(\sum_{j' \le j} q_{j'}\right)\right) p_{i^*} \left(q_{i^*} + q_{i}\right).
\end{align*}
{
Equivalently, we only need to show
\begin{align*}
     \left(\sum_{j = i^*}^{i - 1} t_j p_j q_i \right) q_{i^*} \le \left(\sum_{j = 1}^k t_j p_j \left(\sum_{j' \le j} q_{j'}\right)\right)q_{i},
\end{align*}
where the equivalence follows from canceling the common term $\left(\sum_{j = 1}^k t_j p_j \sum_{j' \le j} q_{j'} \right) \left(p_{i^*} q_{i^*}\right)$ in the sum. 
We note that $q_{i*} \le \sum_{j' \le j} q_{j'} $ for any $j \ge i^*$, 
the inequality above is always true since
\begin{align*}
    \left(\sum_{j = i^*}^{i - 1} t_j p_j \right) q_{i^*} \le \left(\sum_{j = i^*}^{i - 1} t_j p_j \sum_{j' \le j} q_{j'}\right) \le \sum_{j = 1}^k t_j p_j \left(\sum_{j' \le j} q_{j'}\right).
\end{align*} 
Therefore, we conclude that the inner min is always obtained at some $q_{i^*} = 1$.

Since the inner min is always obtained at some $q_{i^*} = 1$, the maximization of ${\rm CR}(X, \mathcal{F})$ will be 
\begin{align*}
    & \max_{t_1,\dots,t_k} \min_{i \in [k]} \quad  \left\{ \frac{\sum_{j \in [k]: j \ge i} t_j p_j}{p_i}\right\}\\
    & \text{subject to } \, \quad \sum_{j} t_j = 1, \quad t_j \ge 0 \quad \forall j \in [k].
\end{align*}
which is equivalent to \cref{eq:P2_reformed} (P2). }
By the result of \cref{eq:P2_reformed} (P2), we know the ${\rm CR}(X, \mathcal{F})$ is at most $1/k$. Further, the ratio $1/k$ is tight, when $X$ is the optimal solution of \cref{eq:P2_reformed} (P2), the competitive ratio is exactly $1/k$.
\hfill\Halmos\endproof


\section{Omitted proofs from Section \ref{sec:adap}} \label{app:adap_proofs}


\subsection{Proof of \cref{lem:unbiased_estimates}}

We use induction on $j \leq k$.  For the base case $j=1$, recall that $D_1 \sim {\rm Binomial}(n_1 ,q(s_1))$.  
Thus, \[\E[ \hat n_1] = \frac{\E[D_1]}{q(s_1)}  = n_1.\]

Now consider $j$ such that $1 < j \le k$. As the \ih, we assume that $\widehat n_i$ is unbiased for any $i < j$.  
Note that \[D_i \sim {\rm Bin}\lb(\sum_{i \leq j}n_i - \sum_{i < j} D_i,\ q(s_j)\rb),\] so 
\[ \begin{split}
\E\lb[\widehat n_j \mid \{D_i\}_{i < j}\rb] &= \frac 1{q(s_j)}\E[D_j \mid \{D_i\}_{ < j} ] - \sum_{i < j}(\widehat n_i - D_i) \\
&= \sum_{i\leq j} n_i - \sum_{i < j} D_i - \sum_{i < j}(\widehat n_i - D_i) \\
&= n_j + \sum_{i < j}(n_i - \widehat n_i)
\end{split} \]
By the inductive hypothesis, we conclude that
\[ \E[ \widehat n_j] =\E[\E[\widehat n_j \mid \{D_i\}_{ i< j}]] = n_j + \sum_{i < j} \E[n_i - \widehat n_i] = n_j. \eqno \Halmos\]


\subsection{Proof of \cref{lem:regret_decomp_two_price}}
The revenue of our policy can be broken up into the revenue during the learning phase and the revenue during the earning phase.  The revenue during the earning phase is ${\rm Rev}( (1-s)\widehat {\bf t}, {\cal I}_{\rm rem})$, which is thus a lower bound on the total revenue of our policy.  
Writing down the regret of our policy on instance $\cI$, we have
\begin{align*}
    {\rm Regret}(X^{\rm LTE}, \cI) \leq  \opt(\cI) - \E[{\rm Rev}( (1-s)\widehat {\bf t}, {\cal I}_{\rm rem})].
\end{align*}
Expanding this further, we have
\begin{align*}
\opt(\cI) - \E[{\rm Rev}( (1-s)\widehat {\bf t}, {\cal I}_{\rm rem})] = \E[{\rm Rev}(\widehat {\bf t}, \cI)] - \E[{\rm Rev}( (1-s)\widehat {\bf t}, {\cal I}_{\rm rem})]
+ \opt(\cI) - \E[{\rm Rev}(\widehat {\bf t}, \cI)],    
\end{align*}
which is at most $\eta_1 + \eta_2$, completing the proof.
\hfill\Halmos

\subsection{Proof of \cref{lem:k_price_eta1_bound} (and \cref{lem:eta1_bound})}
For a non-adaptive policy ${\bf t} = (t_1,t_2,\ldots, t_k)$  
We prove this lemma in two steps.  
First, we account for the fact that $(1-\sum_{j=1}^k s_j){\bf t}$ operates on a shorter time horizon than ${\bf t}$, then we account for the demand that was lost during exploration.  

For the first step, note that for any $j \in [k]$ and any non-adaptive policy ${\bf t}$, we have
\[
r_j({\bf t}) \ge r_j\lb(\lb(1-\sum_{j=1}^k s_j \rb)  {\bf t} \rb) \ge \lb(1-\sum_{j=1}^k s_j \rb) r_j({\bf t}).
\]
The first inequality follows since ${\bf t}$ is  the same as $\lb(1-\sum_{j=1}^k s_j \rb){\bf t}$, but operates on a longer time horizon.  The second comes from the following inequality which follows from convexity of $\exp(\cdot)$. 
Note that for all $x \in \mathbb{R}$ and $\alpha \in (0,1)$, we have $1- \exp(-\alpha x) \geq \alpha(1-\exp(-x)),$ so
\[ \E[ {\rm Rev}( (1-s)\widehat {\bf t},\ {\cal I}_{\rm rem})]\ge \E[ {\rm Rev}( \widehat {\bf t},\ {\cal I}_{\rm rem})] - \E[{\rm Rev}( \widehat {\bf t},\ {\cal I}_{\rm rem})] \cdot \sum_{j=1}^k s_j \ge \E[ {\rm Rev}( \widehat {\bf t},\ {\cal I}_{\rm rem})] - \lambda p_1n\sum_{j=1}^k s_j,
\]
where the last inequality is because $\E[{\rm Rev}( \widehat {\bf t},\ {\cal I}_{\rm rem})] \le \lambda p_1 n$.

Next we account for the difference between $\cI_{\rm rem}$ and $\cI$ due to the learning phase.
Let $D^j_i$ be the number of type-$j$ customers who purchase in the $i$'th exploration period (this may be 0) so that $\sum_{j=1}^k D^j_i = D_i$.
In the instance $\cI_{\rm rem}$, we have that $N_j = n_j - \sum_{i=1}^k D^j_i$.
Now observe that for any non-adaptive policy ${\bf t}$ (which may depend on the random instance $\cI_{\rm rem}$), we have the following bound that holds for any sample path.
\[
    {\rm Rev}({\bf t}, {\cal I}) - {\rm Rev}({\bf t}, \cI_{\rm rem}) = \sum_{j=1}^k r_j({\bf t}) (n_j - N_j) = \sum_{j=1}^k r_j({\bf t}) \sum_{i=1}^kD^j_i \leq p_1 \sum_{i=1}^k D_i
\]
The last inequality follows by observing that $r_j(X) \leq p_1$ and switching the order of summations.
Thus, 
\[\mathbb E\lb[{\rm Rev}\lb(\widehat {\bf t},\ \cI_{\rm rem}\rb)\rb] \ge \mathbb E\lb[{\rm Rev}\lb(\widehat {\bf t},\ {\cal I}\rb)\rb] - p_1 \sum_{i=1}^k \mathbb E[D_i].\]
Thus,
$$\mathbb E[D_i \mid (D_j)_{j < i}] = q(s_i) \left( \sum_{j \leq i} n_j - \sum_{j < i} D_j\right) \leq \lambda s_i n,$$
so summing over all $i$, applying the tower rule and the above inequality completes the proof.
\hfill\Halmos\endproof

\subsection{Proof of \cref{lem:eta2_bound}}
We recall the basic version of Bernstein's inequality, which in our case will give us slightly tighter concentration than if we applied Hoeffding's inequality.

\begin{theorem}[Bernstein's Inequality]\label{thm:concentration} Let $X_1,\dots,X_n$ be i.i.d. bounded between $[-1,1]$, each with mean $p$ and variance $\sigma^2$. 
Then, for any $\Delta>0$, \[\ho{P}\lb[\lb|\sum_{i=1}^n X_i - np\rb| > \Delta \rb] \le \exp\lb(-\frac {\frac 12 \Delta^2}{n\sigma^2 + \frac 13\Delta}\rb). \]
\end{theorem}

As standard in the MAB literature, we will consider the event where the estimation \sats\ the \ci\ given by concentration bounds. 


\begin{definition}[{\bf Clean Event}]
Define the \emph{clean event}\[{\cal E} = \lb\{\lb|\widehat n_i - n_i\rb| \le \sqrt{\frac{8n \log n}{\lambda s}},\quad \forall i=1,2\rb\}.\]
\end{definition}

\begin{lemma}[Clean Event is Likely] \label{lem:clean_event}
    
We have $\ho{P}[{\cal E}] \geq 1- 2n^{-2}$ whenever $\lambda s\leq 1.59$.
\end{lemma}
\proof{Proof.}
Recall that $\hat n_1 = D_1/q(s)$, where $D_1 \sim {\rm Binomial}(n_1, q(s))$.
{\new Applying Theorem~\ref{thm:concentration} to $D$ and note that $\sigma^2= q(1-q)\le q$ we have \begin{align}\label{eqn:122223}
\ho{P}\lb[|\hat n - n| \ge \delta n \rb] = \ho{P}\lb[\lb|D_1 - nq\rb| \ge q\delta n \rb] \le \exp\lb(-\frac{\frac 12(q\delta n)^2}{n\sigma^2 + \frac 13 q\delta n}\rb)\le \exp\lb(-\frac{q n \delta ^2}{2\lb(1+\frac \delta 3\rb)}\rb).
\end{align}
Taking $\delta = \sqrt{\frac{8\log n}{nq}}$, then  
\[\ho{P}\lb[\lb|n_1- \hat n_1\rb | \le \sqrt{\frac{8n\log n}{\lam s}}\rb] \le \eqref{eqn:122223} = \ho{P}\lb[\lb|\hat n_1 - n_1\rb|\ge \sqrt{\frac{8n\log n} q}\rb] \le n^{-2}.\]
}
Since $n_2 = n - n_1$ and $\hat n_2 = n - \hat n_1$, we have $|\hat n_2 - n_2| = |\hat n_1 - n_1|$, and  therefore by union bound, we have 
\[\ho{P}[\mathcal{E}] \ge  1- 2n^{-2}.\eqno \Halmos\]


The following lemma is standard and states that if two functions are point-wise close, then
so are their maximums.

\begin{lemma}[Similar Functions Have Similar Maximum Values]
\label{lem:fg_lemma}
Let $f,g$ be any functions defined on any set $\cX$. 
If for all $x \in \cX$, we have $|f(x) - g(x)| \leq \epsilon$, then $|\max_x f(x) -\max_x g(x)| \leq 3 \epsilon$.
\end{lemma}
\proof{Proof.}  \Sps\ $x^*_f \in \arg\max_x f(x)$ and $x^*_g \in \arg\max_x g(x)$.  
By symmetry, it suffices to show $f(x^*_f) - g(x^*_g) \le 3 \epsilon$. Expanding this difference, we have
\[ f(x^*_f) - g(x^*_g) = f(x^*_f) - g(x^*_f) + g(x^*_f)  -f(x^*_g) + f(x^*_g) - g(x^*_g) \leq 2\eps + g(x^*_f) - f(x^*_g) \leq 3 \eps,\]
where the second follows since $g(x^*_f) - f(x^*_g) \leq g(x^*_g) - f(x^*_g) \leq \epsilon$.
\hfill\Halmos\endproof


\noindent{\bf Proof of \cref{lem:eta2_bound}.}
We aim to bound $|\E[{\rm Rev}(\widehat {\bf t}, \cI)] - \opt(\cI)|$.  First, we observe that $\opt(\cI) = {\rm Rev}({\bf t}^*, \cI)$ for some optimal non-adaptive policy ${\bf t}^*$.  Thus, under any sample path we have:
\begin{align*}
    |{\rm Rev}(\widehat {\bf t}, \cI) - \opt(\cI)| & \leq |{\rm Rev}(\widehat {\bf t}, \cI) - {\rm Rev}(\widehat {\bf t}, \widehat\cI)|  + |{\rm Rev}(\widehat {\bf t}, \widehat\cI) - {\rm Rev}({\bf t}^*, \cI)| \\
    & \leq |R_1(\widehat {\bf t})(n_1 - \widehat n_1) + R_2(\widehat {\bf t})(n_2 - \widehat n_2)| + 3 \sup_{\bf t} |{\rm Rev}( {\bf t}, \widehat\cI) - {\rm Rev}\lb({\bf t}, \cI\rb)| \\
    & \leq 5 p_1(|\widehat n_1 - n_1| + |\widehat n_2 - n_2|).
\end{align*}

The first step is the triangle inequality.  The second follows from the definition of ${\rm Rev}(\cdot)$ and \cref{lem:fg_lemma}.  The last inequality again follows from the definition of ${\rm Rev}(\cdot)$ and the triangle inequality.

Thus we may bound $\eta_2$ under the two conditions ${\neg \cal E}$ and ${\cal E}$, where ${\cal E}$ is the event defined in Lemma~\ref{lem:clean_event}.
For the first case, this difference is at most $np_1$ and $\ho{P}[\neg {\cal E}] \leq 2n^{-2}$, so the expected contribution for this case is $o(1)$.  
For the second case, we use the inequality above that bounds $\eta_2$ in terms of the estimation error and apply the bound on the estimation error under event ${\cal E}$.
Combining the bounds under each event yields the lemma.
\hfill\Halmos\endproof

\section{Regret Analysis for the $k$-Price Setting}
We proved the bound for $\eta_1$ in the $k$-price setting above, so here we focus on bounding $\eta_2$.

\subsection{Proof of \cref{thm:mgf_bound}}\label{apdx:mgf_bd} 
We proceed by induction on $k$.  The base case (i.e., $k=1$) holds trivially. In fact, by the definition of conditional subgaussianity, we have
\[ \E\lb[e^{\gamma r_1 \Delta_1}\rb] \le e^{(\gamma r_1)^2c_1^2/2}.\]

Now \sps\ the claim holds for $k-1$.
For any \xulie\ ${\bf a} = (a_1,\dots,a_k)$ and $\ell \le k$, let us write the $\ell$-prefix as ${\bf a}^\ell = (a_1,\dots ,a_\ell)$.
Then, by the law of total \prb, we have 
\begin{align}\label{eqn:022824c}
\E\lb[e^{\gamma \lan {\bf r}, {\bf \Delta}\ran}\rb] = \E\lb[e^{\gamma r_k \Delta_k} \cdot e^{\gamma \lan {\bf r}^{k-1}, {\bf \Delta}^{k-1} \ran}\rb] 
= \E\lb[ \E\lb[ e^{\gamma r_k \Delta_k} \mid {\cal F}_{k-1}\rb] \cdot e^{\gamma \lan {\bf r}^{k-1}, {\bf \Delta}^{k-1} \ran}\rb].
\end{align}
Note that $\bf \Delta$ is self-neutralizing and conditionally sub-gaussian, and that $r_k$ is a constant, we can bound the conditional expectation as  
\[\E\lb[ e^{\gamma r_k \Delta_k} \mid {\cal F}_{k-1}\rb] 
\le \exp \lb(\gamma r_k \cdot \E[\Delta_k \mid {\cal F}_{k-1}] +  \frac 12 \gamma^2 (r_kc_k)^2\rb) \leq \exp \lb(-\gamma r_k \sum_{j=1}^{k-1} \Delta_j + \frac 12\gamma^2 (r_k c_k)^2 \rb).\]
Applying the above to \cref{eqn:022824c}, and factoring out the deterministic part, we obtain
\begin{align}\label{eqn:022824d}
\E\lb[e^{\gamma \lan {\bf r}, {\bf \Delta}\ran}\rb] 
&\le e^{\frac 12 \gamma^2 (r_kc_k)^2} \cdot \E\lb[\exp\lb(\gamma  \lan {\bf r}^{k-1}, {\bf \Delta}^{k-1}\ran  - \gamma r_k \sum_{j=1}^{k-1} \Delta_j \rb)\rb]\notag\\
& = e^{\frac 12 \gamma^2 (r_k c_k)^2} \cdot \E\lb[e^{\gamma \lan {\bf r'}, {\bf \Delta}^{k-1}\ran}\rb].
\end{align}
where ${\bf r'}=(r_j')_{j\in [k-1]}$ is given by $r_j' = r_j - r_k$.
Note that ${\bf r'}$ is still non-negative and non-increasing, so the \ih\ implies that
\[ \E\lb[e^{\gamma \lan {\bf r'}, {\bf \Delta^{k-1}}\ran}\rb] \leq \exp \lb(\gamma^2 \sum_{j=1}^{k-1} (r_j' c_j)^2/2 \rb) \leq \exp \lb(\gamma^2 \sum_{j=1}^{k-1} (r_j c_j)^2/2 \rb)\]
The induction step follows immediately by combining this with \cref{eqn:022824d}.
\hfill\Halmos\endproof

\subsection{Verifying the Two Properties of the Error Process}\label{apdx:error_process}
To apply the mgf bound in \cref{thm:mgf_bound} on the error process ${\bf \hat n-n}$, we need to verify that this process is self-neutralizing and conditionally subgaussian.

\begin{proposition}
[$\widehat n_j - n_j$ is self-neutralizing] \label{lem:errors_self_neutralizing}
    We have $\E[\widehat n_j - n_j \mid {\cal F}_{j-1}] = - \sum_{i=1}^{j-1} (\widehat n_i - n_i)$ for each $j \in [k]$.
\end{proposition}

\proof{Proof.} From the definition of $\widehat n_j$, we have
\[{\mathbb E}[(\widehat n_j - n_j) \mid {\cal F}_{j-1}] =   \frac{{\mathbb E}[D_j \mid {\cal F}_{j-1}]}{ q(s_j)} - \sum_{i \leq j-1}(\widehat n_i - D_i) - n_j = -\sum_{i \leq j-1}(\widehat n_i - n_i),\]
where the second equality follows since $D_j \sim {\rm Bin}\lb(\sum_{i\leq j} n_i - \sum_{i \leq j-1} D_i,\ q(s_j)\rb)$, and $D_i$'s cancel each other.
\hfill\Halmos 

\begin{proposition}[$\widehat n_j - n_j$ is conditionally sub-gaussian] \label{lem:errors_conditionally_subgaussian}
\Sps\ $0< \gamma \le 1.79 q(s_j)$. Then, for each $j \in [k]$ we have 
\[ \E\lb[e^{\gamma (\widehat n_j - n_j)} \mid {\cal F}_{j-1}\rb] \le \exp\lb(- \gamma \sum_{i=1}^{j-1} (\widehat n_j - n_j) + \gamma^2 \sum_{i \leq j} n_i /q(s_j) \rb).\]
\end{proposition}
\proof{Proof.} By the definition of $\widehat n_j$, we have
\begin{align*}
\E\lb[e^{\gamma (\widehat n_j - n_j)} \mid {\cal F}_{j-1}\rb] &= \E\lb[ \exp\lb(\gamma \lb(\frac {D_j}{q(s_j)} - \sum_{i \leq j-1} (\widehat n_i - D_i) - n_i\rb)\rb) \mid {\cal F}_{j-1} \rb] \\
& = \exp\lb(-\gamma \sum_{i \leq j-1} (\widehat n_i - D_i) -n_i \rb) \cdot \E\lb[\exp\lb(\frac{\gamma}{q(s_j)}D_j\rb) \mid {\cal F}_{j-1} \rb]
\end{align*}
To bound the latter term, recall that \[D_j \sim {\rm Bin}\lb(\sum_{i\leq j} n_i - \sum_{i \leq j-1} D_i,\ q(s_j)\rb).\]
Writing $N_j = \sum_{i \leq j} n_j - \sum_{i \leq j-1} D_i$, we obtain
\[\E\lb[\exp\lb(\frac{\gamma}{q(s_j)}D_j\rb) \mid {\cal F}_{j-1} \rb] = (1 + q(s_j)(\exp(\gamma/q(s_j)) -1))^{N_j} \leq \exp(q(s_j) N_j(\exp(\gamma /q(s_j)) - 1)).\]
The first step uses the moment generating function of a binomial distribution, and the second step applies $1+x \leq e^x$.
Note that $e^x \leq 1+ x + x^2$ for $x \le 1.79$, so we deduce that
\[ \E\lb[\exp\lb(\frac{\gamma}{q(s_j)}D_j\rb) \mid {\cal F}_{j-1} \rb] 
\le \exp\lb(\gamma N_j + \gamma^2 \frac{N_j}{q(s_j)} \rb)\le \exp \lb(\gamma N_j +\gamma^2 \sum_{i \leq j} \frac{n_i}{q(s_j)}\rb),\]
where the last step follows since $N_j \leq \sum_{i=1}^j n_i$ a.s. 
Plugging the above bound and simplifying yields the lemma.
\hfill\Halmos



\subsection{Proof of \cref{lem:convex_comb_argument}}
For any non-increasing $r\in \real^k_+$, we have 
\begin{align}\label{eqn:022824g}
r = r_k \phi_k + \sum_{j=1}^{k-1} (r_j - r_{j+1}) \phi_j + (1-r_1)\phi_0.
\end{align}
Note that for each $j\in [k]$ we have $r_j \geq r_{j+1}$ and \[r_k + \sum_{j=1}^{k-1} (r_j - r_{j+1})  + 1-r_1 = 1,\] 
and therefore \cref{eqn:022824g} is a valid \cvx\ combination.
\hfill\Halmos

\subsection{Proof of \cref{lem:k_price_eta2_bound}}


Note that the policy space is the closed simplex, which is a compact set, and that the revenue function of any UDPM instance is \conti, so $\sup_{\bf t} {\rm Rev}\lb({\bf t}, \cI\rb)$ can be attained, say, at $\bf t^\star$.
Then, by the definition of $\eta_2$, we have  
\begin{align*}
\eta_2 
&= \lb|\E \lb[{\rm Rev}\lb(\widehat {\bf t}, \cI\rb)\rb] - \E[{\rm Rev}\lb({\bf t}^\star, \cI\rb)]\rb|\\
& \le 4\cdot \E\lb[\sup_{{\bf t}\in \Pi({\cal P})} \lb|{\rm Rev}\lb({\bf t}, \widehat \cI\rb) - {\rm Rev}\lb({\bf t}, \cI\rb)\rb|\rb] \\
&= 4 \int_{0}^{\infty} \ho{P}\lb[ \sup_{{\bf t}\in \Pi({\cal P})} |{\rm Rev}\lb({\bf t}, \widehat \cI\rb) - {\rm Rev}\lb({\bf t}, \cI\rb)| >\tau\rb]\ d\tau \\
& \le 4\left( \tau + p_1 n\cdot \ho{P}\lb[ \sup_{{\bf t}\in \Pi({\cal P})} |R({\bf t}, \widehat \cI) - R({\bf t}, \cI)| > \tau\rb]\right).
\end{align*}
The first inequality follows from the triangle inequality and the second from Jensen's inequality.  
The third follows from \cref{lem:fg_lemma}.  
The last inequality follows since the absolute difference is at most $p_1 n$. \hfill\Halmos